\documentclass[11pt]{article}
\usepackage{titlesec}
\usepackage{titletoc}
\usepackage[title,titletoc]{appendix}
\usepackage[hyphenbreaks]{breakurl}
\usepackage[hyphens]{url}
\usepackage{bm}
\usepackage{authblk}
\usepackage{amsmath}
\usepackage{algorithm}
\usepackage[noend]{algpseudocode}
\usepackage{hyperref}
\usepackage{url}
\usepackage{graphicx}
\usepackage{float}
\usepackage{subfigure}
\usepackage{bm}
\usepackage{amssymb}
\usepackage{footmisc}
\usepackage{enumitem}
\usepackage{lipsum}
\usepackage{upgreek}
\algdef{SE}[DOWHILE]{Do}{doWhile}{\algorithmicdo}[1]{\algorithmicwhile\ #1}%
\usepackage[top=1in, bottom=1.25in, left=1.1in, right=1.1in]{geometry}

\makeatletter
\def\BState{\State\hskip-\ALG@thistlm}
\makeatother

\makeatletter
\newcommand{\Spvek}[2][r]{%
  \gdef\@VORNE{1}
  \left(\hskip-\arraycolsep%
    \begin{array}{#1}\vekSp@lten{#2}\end{array}%
  \hskip-\arraycolsep\right)}

\def\vekSp@lten#1{\xvekSp@lten#1;vekL@stLine;}
\def\vekL@stLine{vekL@stLine}
\def\xvekSp@lten#1;{\def\temp{#1}%
  \ifx\temp\vekL@stLine
  \else
    \ifnum\@VORNE=1\gdef\@VORNE{0}
    \else\@arraycr\fi%
    #1%
    \expandafter\xvekSp@lten
  \fi}
\makeatother

\begin{document}
\title{\textbf{Object Manipulation Learning by Imitation}}
\author{Zhen Zeng, Benjamin Kuipers
\\Electrical Engineering and Computer Science, University of Michigan
\\zengzhen@umich.edu, kuipers@umich.edu}
\date{Mar 15, 2016}
\maketitle

\begin{abstract}
We aim to enable robot to learn object manipulation by imitation. Given external observations of demonstrations on object manipulations, we believe that two underlying problems to address in learning by imitation is 1) segment a given demonstration into skills that can be individually learned and reused, and 2) formulate the correct RL (Reinforcement Learning) problem that only considers the relevant aspects of each skill so that the policy for each skill can be effectively learned. Previous works made certain progress in this direction, but none has taken private information into account. The public information is the information that is available in the external observations of demonstration, and the private information is the information that are only available to the agent that executes the actions, such as tactile sensations. Our contribution is that we provide a method for the robot to automatically segment the demonstration of object manipulations into multiple skills, and formulate the correct RL problem for each skill, and automatically decide whether the private information is an important aspect of each skill based on interaction with the world. Our experiment shows that our robot learns to pick up a block, and stack it onto another block by imitating an observed demonstration. The evaluation is based on 1) whether the demonstration is reasonably segmented, 2) whether the correct RL problems are formulated, 3) and whether a good policy is learned.
\end{abstract}

\section{Introduction}

\subsection{Why Imitation Learning?}
When a robot is presented with an unfamiliar object, the robot is not aware of what actions can cause what changes to the state of the object. Learning by imitation is an effective way for a robot to gain knowledge about possible useful actions on objects in its environment.

For a robot to learn to effectively manipulate an unfamiliar object to achieve a particular task, there are two problems, a) how the robot knows what the task is: for people familiar with programming, they can directly put in the corresponding criteria to define the task for the robot, but it is not realistic for most of the consumers of commercial robots who are usually not familiar with programming with robot, b) how the robot learns to achieve the task: it is difficult for the robot to automatically learn to achieve the task in a short amount of time just based on pure self-exploration. 
One natural way to inform the robot about a task is to show it how an expert completes the task. The robot should be able to represent the desired manipulation behavior in a way such that it can effectively understands "what happened when". And the representation could guide the robot to learn actions to imitate the behavior, thus reproducing the same manipulation effects.

Given an observed behavior, usually formally defined as state variables describing the observed environment and action variables describing the observed actions, the robot should be able to learn a policy, which is a mapping between states and actions, such that it can select an action to execute based on its current state.

\subsection{RL Problem Formulation - MDP}
RL methods are popular for policy learning, and the robot can improve its performance overtime through interaction with the world, thus we use RL methods in our work. To learn a policy with RL methods, we should first define a RL problem, usually formulated as an MDP (Markov Decision Process) with following components: the state space $\mathcal{S}$, the action space $\mathcal{A}$, and the reward function $R(s)$. We'll elaborate on the importance of correctly defining these three components in an MDP. 

\subsubsection{Importance of Reward Function $R(s)$}
For a robot to learn to effectively manipulate an unfamiliar object by imitating other's behavior, it is important for the robot to understand what to imitate, i.e., the goal of the observed behavior. For robotic experts, they can directly hand code the corresponding criteria into $R(s)$ to define the goal for the robot, but it is not realistic for most of the consumers of commercial robots to do so.  
 
Thus it is important for the robot to automatically capture the goal of an observed behavior by defining the correct reward function. For example, if the goal of an observed behavior is to reach and grasp a cup, then the reward function can be defined as -1 everywhere except for a big positive reward when then cup is being grasped. 

\subsubsection{Importance of State, Action Space $\mathcal{S,A}$}
Think of the way we act in our daily lives, when we manipulate some objects of interest, we don't pay much attention to other objects in the environment. For example, if there are two blocks and a cup on a table, when we are stacking the two blocks, we don't really care about where the cup is; on the other hand, when we try to put one of the block into the cup, we don't care where the other block is. This reflects that even we have access to a bunch of information, we always abstract out the most important information relevant to our task by hand, and this is formally referred as \emph{abstraction} in machine learning. Abstraction is a key to select important aspects of a skill, such that the skill is generalized to scenarios where the complete $\mathcal{S,A}$ change. With the correct abstraction, the appropriate state and action space is defined for the RL problems.

\subsubsection{Importance of Segmenting an Observed Behavior into Multiple MDPs}
The observed behavior can underlie multiple policies, i.e., there are several skills involved in the behavior. For example, picking up a block, and then stacking it onto another one are two different skills. When the observed behavior contains multiple skills, the robot should be able to automatically segment the behavior into multiple pieces, with each corresponding to a skill, and formulate individual RL problem for each of them.

The advantage of segmenting an observed behavior into a sequence of skills and learn the policy for each skill is that: a) learned individual skills can be reused in other tasks; b) individual RL problem can be formulated with skill specific state and action space, instead of the overall state and action space. 

\subsubsection{Importance of Acknowledging Private Information}
Given an externally observed behavior, private information such as tactile sensations are not available to the robot. If human demonstrates to rotate a jar's lid to open it, the observed behavior will be the hand rotating together with the jar's lid until it is open. With lack of private information, the robot cannot decide whether the private information is important or not by just observing the behavior. Once the robot understands the goal of the behavior, and selects the correct state abstraction, i.e. the orientations of the hand and lid w.r.t the frame of the jar's body in this case, appropriate RL methods can be applied to learn a policy by temporarily assuming that the private information does not matter. As the robot is just grasping the jar's lid without controlling the holding force, the robot could fail to learn a policy to reliably open the jar's lid. In our work, when the robot encounters failure at learning a good policy to reproduce the observed behavior, it will propose to expand the state space in the originally formulated RL problem to include the private information, i.e., tactile sensation, and expand the action space to include the controller on holding force, then restart the learning based on the not-so-good policy learned previously in the originally formulated RL problem. 

\subsection{Our Contribution}
Given an observed behavior, it is important to segment it into multiple pieces (if there are multiple skills involved), and formulate the correct RL problem for each piece such that RL methods can be further applied to learn policies. Our contribution is two-fold: 
\begin{itemize}
{\setlength\itemindent{4pt} \item 
We provide a method to online select the appropriate abstraction such that 1) the correct reference frame is selected, and 2) the relevant objects are selected.}
{\setlength\itemindent{4pt} \item
When private information is not available in the observed behavior, the robot is able to decide whether the private information is important, and reformulate the RL problems if needed. }
\end{itemize}

Next, Sec \ref{sec:related_work} will discuss some related works, Sec \ref{sec:notations} introduces notations used in this paper, and then Sec \ref{sec:overview}, \ref{sec:segmentation}, \ref{sec:policy_learning}, \ref{sec:reformulation} explains the methods involved in our framework, and Sec \ref{sec:experiment} discuss our experiment setup, and the evaluation results.

\section{Related Work}\label{sec:related_work}
\subsection{Understanding Manipulation Behavior}
A potentially useful manipulation behavior representation
for artificial agents should satisfy several criteria. As pointed out in \cite{aksoy2011learning}, the representation needs to be based on sensory signals and learnable by observation. From the point of view of learning by imitation, the representation should
also satisfy that (1) it is not redundant in the sense that it
should not encode information that exists only within certain observed behaviors, such as specific motion trajectories of human arm joints and objects; (2) it should be simple such that it can be easily interpreted as a roadmap that guide an agent to act.

Previous works on human manipulation actions recognition have attempted to represent manipulation behavior in a probabilistic manner. Human poses, human-object context, and object-object context have been considered to solve the problem jointly as in \cite{koppula2013learning}. Similarily in works by Kjellstrom et al.\cite{kjellstrom2008simultaneous}, pre-defined hand-object features
and manipulation features are extracted, and the semantic
manipulation action-object dependencies are learned based on CRFs.
Their representation of manipulation behavior is powerful in that they can recognize manipulation actions as well as object
categories. Although previous works are robust in manipulation action recognition when presented by various view points and even occlusions, those representations of the behavior cannot be directly translated into any step by step guidance, or a roadmap that an agent can follow for effective imitation.

Works by Aksoy's group \cite{aksoy2011learning} revealed an effective way to represent manipulation behavior. By observing different human demonstrations of the same manipulation behavior, they discovered that the manipulator, i.e. hand, and objects movement trajectory may vary, but there are certain moments that the spatial relations between the hand and objects are similar or identical across all the demonstrations, and these moments are referred as decisive moments. They introduced Semantic Event Chain (SEC) as a novel, generic representation of manipulation behavior, where they encoded the spatial relations between the manipulator and objects only at decisive moments. SEC describes object relations presented in 2D images, where "touching" is defined as the 2D regions of two objects are side by side, and "overlapping" is defined as the 2D regions of two objects overlapping with each other. In our work, instead of describing object positions in 2D terms, we describe an observed behavior with 3D poses of the manipulator and objects in the workspace.

\subsection{Learning by Imitation}
There are different ways of recording demonstrations for a robot to imitate. Some works \cite{ijspeert2002learning}\cite{nakanishi2004learning} mounted sensors on human body, and recorded the joint angles during the demonstrations to teach humanoid robot drumming and walking patterns. We choose to use an external sensor, a Kinect mounted in front of the robot, to record the human demonstrations, since we don't need precise measurements on the motion or joint angles of the articulated human arm or hand, and it costs much less overhead.

For both ways of recording demonstrations as introduced above, there exists a corresponding issue. The corresponding issue is understood as the identification of a mapping from the demonstrator to the robot that allows the transfer of information. For example, in the case of robot learning to walk by observing human joint angles during demonstrations, before it can imitate the walking pattern, the robot needs to know the mapping between the human joint angles and its own joint angles. For more details on works that address the correspondence issue, please refer to this survey\cite{argall2009survey}. In our case, the robot already knows the hand in the demonstration is correspondent to its grippers since only the hand and the grippers are the actors in the manipulation actions.

In domains like learning biped walking, and dancing patterns, previous works learn robot movements by imitating joint trajectories \cite{ijspeert2002learning}\cite{nakanishi2004learning}\cite{ijspeert2002movement}. They mounted sensors on human body, and recorded the joint angles during the demonstrations to teach humanoid robot drumming and walking patterns. 

While in the domain of learning object manipulations, where the robot needs to learn a manipulation task, directly imitating exact joints trajectories for manipulation tasks won't generalize well. Two completely different sequences of joints trajectories can be performing the same manipulation task. For example, when we pick up an object, we can approach it from various angles and along various joint trajectories, but all these trajectories correspond the same manipulation task. Thus what really matters in the domain of object manipulation learning is to capture and imitate the important aspects of the behavior rather than imitating the quantitative joints trajectories. For example, the important aspects of 
picking up an object are the hand needs to approach the object first, then grasp it to lift it up, no matter what the joints trajectories are.

\subsubsection{Policy Derivation Approaches}
There are two core approaches for policy derivation from observed behavior: RL methods, and direct mapping function learning. Kohl and Stone \cite{kohl2004policy}\cite{kohl2004machine} parameterized a quadruped walk on their Sony Aibo robot, and the robot effectively learned a fast walk based on a proposed policy gradient reinforcement learning method. There have been some works that directly learn the mapping from the state space to the action space. 
With discrete action space, the problem of learning the mapping is essentially a classification problem. For example, Chernova et al. \cite{chernova2007confidence}\cite{chernova2009interactive} learns to navigate through corridors by observing the behavior generated by expert teleoperation. In their work, the states are continuous variables describing the distances of the closets walls, and the actions are discrete variables corresponding to pre-defined controllers that drives the robot forward, to the left, to the right, and u-turn. The mapping from state to action is learned based on GMM. Similar works have been focused on high-level primitive actions such as hand gestures, for learning box and ball sorting tasks \cite{rybski1999interactive}\cite{chernova2008teaching}. With continuous action space, the problem of learning the mapping is then a regression problem. Grollman et al. \cite{grollman2007learning} has applied locally weighted projection regression to soccer skill learning task on an AIBO robot.

Among these two core approaches to policy derivation, RL approach is the only one that can update the policy to potentially improve the behavior beyond the observed one. Furthermore, with the option framework\cite{sutton1999between} that hierarchically combines multiple policies together, RL approach can deal with not only low level control, but also support high level planning. Thus we are pursuing the RL approach as it is the most promising one.

\subsubsection{Learning of MDP Components}
Some previous works have been focused on learning the state space given the reward function(or rewards received from the external environment). Given the reward function, Konidaris and Barto\cite{konidaris2008sensorimotor}\cite{konidaris2009efficient} selects a state abstraction from a set of hypothesized ones based on value function approximation errors. Li et al.\cite{li2006towards} starts with a large state space, then compresses it with variable removal or state aggregation, while it could be infeasibly difficult to learn a value function for the beginning large state space. McCallum\cite{mccallum1996learning} provides an alternative approach to start with empty state space, and then add in new state variables one at a time when it becomes evident that they are necessary for learning the behavior, but requires a significant amount of data and computation to determine which variable to introduce.

Some other works have been focused on learning the reward function given the state space and action space. Reward function is a very important component in RL problem formulation, since it underlies the goal (and maybe desired properties and constraints) of the behavior. Given the state space and action space, state transition model, and observed behavior (as a sequence of state and action pairs), inverse RL methods aims at recovering the reward function. In Markov decision processes, Ng\cite{ng2000algorithms} and Abbeel\cite{abbeel2004apprenticeship} learns the reward function such that the policy underlies the observed behavior is optimal. In their car driving simulation, given observed behavior of different "driving styles", their method was able to learn the reward functions such that each reward function can lead to the optimal policy that generates behvaior qualitatively similar to the observed ones. Kolter and Abbeel \cite{kolter2007hierarchical} extended the inverse RL paradigm to accept isolated advice at different hierarchical levels of the control task, such that it is still feasible to learn the reward function when it is non-trivial to provide the optimal behavior, such as navigating a quadruped robot over extreme terrain. 

Our work focus on automatically formulating RL problems for the skills observed in demonstration, thus learning all the components of the MDPs, and solving the RL problems result in good policies that can reproduce the observed behavior. One of the important step made in this direction is by Konidaris et al. \cite{konidaris2011robot}, they provided an elegant approach for behavior segmentation, and abstraction selection for formulating the RL problems, but their approach was not readily applicable to the object manipulation domain, as discussed later in our experiment section \ref{sec: vs_konidaris}.
\section{Notations \label{sec:notations}}
First of all, $\mathcal{S}$ is the overall state space, and it divides into the overall public state space ${\mathcal{S}}_{public}$ and the overall private state space ${\mathcal{S}}_{private}$. In ${\mathcal{S}}_{public}$, each dimension is a public state variable whose value can be externally observed, such as the $x$ coordinate of an object center, and a boolean state variable indicating whether the hand is open or not. In ${\mathcal{S}}_{private}$, each dimension is a private state variable whose value cannot be externally observed, such as tactile sensations on the fingers. 
 
The robot observes a sequence of states sampled at the frame rate,
\begin{equation}\label{eq:observation} 
O = ( s_0, s_1, \cdots, s_{final} )
\end{equation}
where $s_i \in {\mathcal{S}}_{public}$. More specifically, $\mathbf{s}_i$ is a vector composed of the pose vectors of objects in workspace, and the pose vector of hand, 
\[ s_i = [ P_i^{o_1}, P_i^{o_2}, \cdots, P_i^{h} ]^T \]
where $P_i^{o_j}=[loc_i^{o_j}, ori_i^{o_j}]$ is the pose vector of the $j$th visible object at $i$th sampled frame, and $P_i^h$ is the pose vector of the hand at $i$th sampled frame. All the pose vectors in the observations are in the world frame, they can be transformed into any object frame or the hand frame given the object or hand pose. 
 
Robot action space $\mathcal{A}$ is composed by end-effector movements (translation and rotation in 3D), open and close gripper with commanded holding force. We formulate the movements of the end-effector of our manipulator robot through a set of Dynamic Movement Primitives(DMPs), as introduced by Schaal\cite{schaal2006dynamic} and Ijspeert\cite{ijspeert2013dynamical}. A DMP represents a parameterized nonlinear dynamical systems which is able to encode both discrete (i.ie., point to point) and rhythmic (periodic) trajectories. The action space in our work is composed by joint torques, whose values are determined by a set of control variables in DMPs, and these control variables are the policy parameters being updated during RL.

In the object manipulation domain, we assume that the relevant state variables of a skill are relation variables among several reference frames, and the skill specific reward $R(s)$ is 0 everywhere except for a large positive reward when the relation variables are within some desired region, as observed at the end of the observed behavior. For example, given observed behaviors of a task which is to place a gripped block onto a marker lying on a table, we should be able to conclude that the relevant states are the relative pose between the reference frame associated with the bottom surface of the block, and the reference frame associated with the marker, and the task specific reward is 0 everywhere except for a large positive reward when the relative pose between those two reference frames is for example 0 in each dimension. This is a simple instance of inverse reinforcement learning methods. These synthesized rewards are reasonable since many tabletop object manipulations involve skills with its own ending goal. If the real reward function is more complex than that, other inverse reinforcement learning methods \cite{abbeel2004apprenticeship} can be applied to infer the reward function from the demonstration. 

Demonstrations on tabletop object manipulations can involve multiple skills, for example, grasp and pick up a cup, the series of actions can be segmented into three individual skills: first, approach the cup; second, grasp the cup; third, lift the cup. Each skill is formally defined as an option $o$, as introduced in \cite{sutton1998between}. An option includes three components: 1) an option policy $\pi_o(s,a)$ which gives the probability of executing each action in each state in which the option is defined; 2) an initiation set indicator function ${\mathcal{I}}_o(s)$ which gives 1 for states where the option can be executed and 0 elsewhere; 3) termination condition $\beta_o(s)$ which gives the probability of option execution terminating in states where it is defined.
 
In our work, we aim to automatically select the appropriate state abstraction $M=\langle O_{rel}, o_{ref} \rangle$ such that the correct object reference $o_{ref} \in \{o_0,o_1,\cdots,o_n \}$, and the relevant objects $O_{rel}$ are selected. Specifically, given an abstraction $M=\langle O_{rel}, o_{ref} \rangle$, the complete state space is mapped to an abstracted state space, $\mathcal{S} \mapsto {\mathcal{S}}_{abs}$, where
\[ {\mathcal{S}}_{abs} = \{P^{o_k o_{ref}} | o_k \in O_{rel} \} \]
where $P^{o_i o_j}$ refers to pose of object $o_i$ relative to object $o_j$, thus ${\mathcal{S}}_{abs}$ is composed by the poses of relevant objects $O_{rel}$ relative to object $o_{ref}$.
 
To formulate the correct RL problem for each segmented skill, the correct state abstraction should be selected, and the private action control variable should be considered when it is important.
\section{Method Overview}\label{sec:overview}
\begin{figure}[h!]
\centering
\centerline{\includegraphics[width=1.3\textwidth]{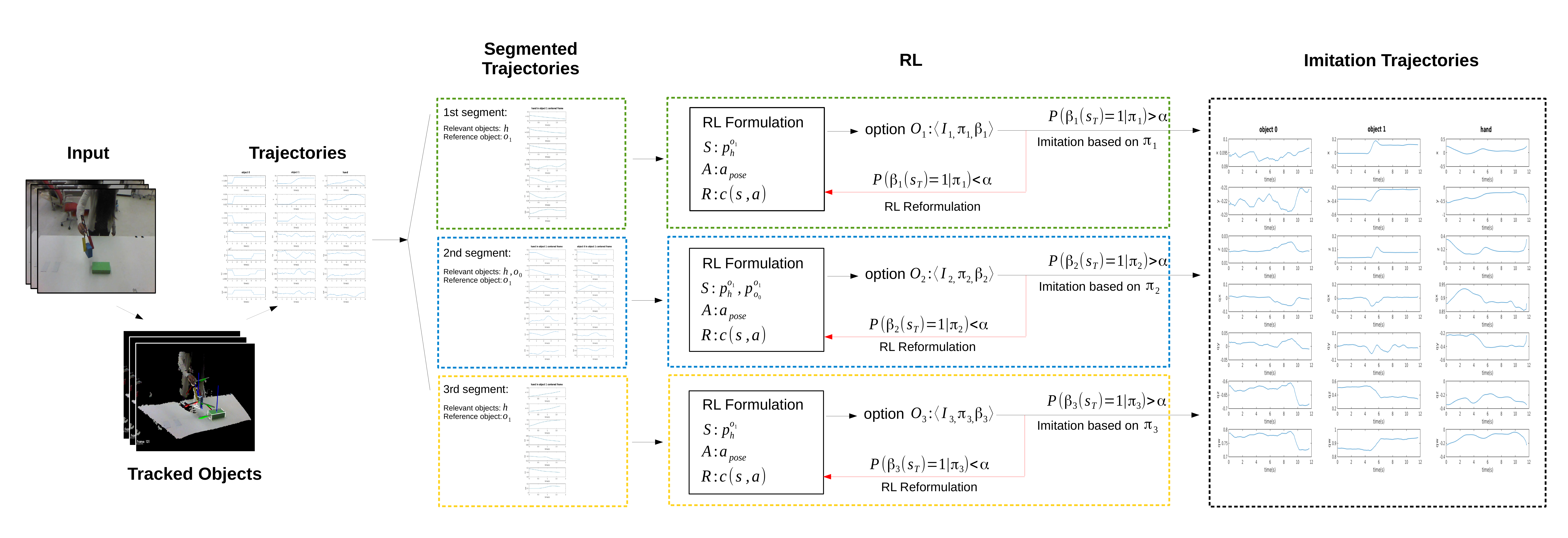}}
\caption{Workflow overview. \textbf{Input:} a series of RGB-D frames that records an observed behavior, e.g., picking up a blue block and stack it onto a green block in the workspace. \textbf{Tracked Objects:} 3D poses of both hand and objects are tracked. \textbf{Trajectories:} pose trajectories of hand and objects in a world reference frame, e.g., tabletop reference frame. Pose variables are $x,y,z,q_x,q_y,q_z,q_w$, where $q$ refers to quaternion. \textbf{Segmented Trajectories:} hand and object pose trajectories are segmented into $K$ pieces, e.g. $K=3$ in the block stacking example. Each segmented behavior is described by pose trajectories of relevant objects in identified reference frame. \textbf{RL:} for each segmented behavior, formulate a RL problem by defining the MDP components $\langle {\mathcal{S}}, {\mathcal{A}}, R \rangle$, and learn the optimal policy $\pi_i$ for each RL problem. which leads to an option $o_i: \langle {\mathcal{I}}_i, \pi_i, \beta_i \rangle$ with an initiation set ${\mathcal{I}}_i \subseteq S$ and termination condition $\beta_i: S \mapsto [0,1]$. The $i$th RL problem is reformulated if $P(\beta_i(\mathbf{s}_T)=1)|\pi_i)$ is below threshold $\alpha$, where $s_T$ is the final state when the trajectory generated by DMPs end. The RL reformulation strategy is explained in sec \ref{sec:reformulation}. \textbf{Imitation trajectories:} once all the learned policies pass the criteria $P(\beta_i(s_T)=1)|\pi_i)>\alpha$, the robot can imitate the observed behavior by following the set of optimal policies sequentially. Note that the imitation trajectories are not generated by directly imitating the observed trajectories, but generated by following the policies learned from RL problems that are automatically formulated by the robot.}
\label{fig:workflow}
\end{figure}

The workflow of our imitation learning method is as illustrated in Figure \ref{fig:workflow}. The robot observes human behavior through an RGB-D camera, e.g., Kinect sensor, and the imitation learning problem is solved by finding the right sequence of RL problem formulations and learning the optimal policy for each RL problem. Our methods consist of following steps: 1) segmenting the observed behavior into pieces with identified relevant objects $O_{rel}$ and reference frame $o_{ref}$; 2) formulate a RL problem for each segmented behavior, and learn the optimal policy for each initially formulated RL problem; 3) reformulate RL problems based on criteria that evaluates whether the imitation following the optimal policy is successful. More detail about our methods are explained in following sections.

\section{Behavior Segmentation}\label{sec:segmentation}
In this section, an observed behavior is automatically segmented in to pieces with simultaneously identified state abstraction $M=\langle O_{rel}, o_{ref} \rangle$, i.e., relevant objects and reference frame. 

Our method is based on the assumption that 1) the origin of the reference object $o_{ref}$ is close to the latent attractor of hand/end-effector trajectory, e.g., the center of a target object serves as the attractor of a reaching trajectory; and 2) when an object $o_i$ is moving along with the manipulator, the object is an relevant object, i.e. $o_i \in O_{rel}$. For example, a block that moves along with the hand when being grasped is an relevant object; 3) we assume the hand is always relevant, i.e., $h \in O_{rel}$. 

Our intuition is that an observed behavior can be segmented at time points where the reference frame changes or relevant objects change. The changepoint detection algorithm \cite{fearnhead2007line} can be applied to find these time points and select the correct abstraction $M$, as explained in following section.

\subsection{Changepoint Detection Algorithm}
Here is a recap on the statistical changepoint detection algorithm in a general regression setting. Given observed data and a set of candidate models $Q$, we assume that the data are sequentially generated by an instance of a single model, occasionally switching to a different model at certain points in time, called changepoints. The goal is to infer the number and positions of the changepoints and select an appropriate model for each segment.

An efficient changepoint detection algorithm was introduced by Fearnhead and Liu \cite{fearnhead2007line} that simultaneously find the MAP changepoints and select model for each segment. Observations are data pairs $(x_t, y_t)$ observed at times $t\in{1,2,\cdots,T}$, and there exists a set of candidate models $Q$ with prior $P(q)$ for each model $q\in Q$. The marginal probability of a segment length $l$ is modeled with probability mass function $g(l)$ and cumulative mass function $G(l)=\sum_{i=1}^l g(l)$. And a data segment from time $j+1$ to $t$ can be fit using model $q$ to obtain $P(j,t,q)$, the probability of the data segment conditioned on $q$. Functions $g(l)$ and $P(j,t,q)$ is either given as prior knowledge or pre-learned based on some training data.

\begin{figure}[h!]
\centering
\centerline{\includegraphics[width=0.6\textwidth]{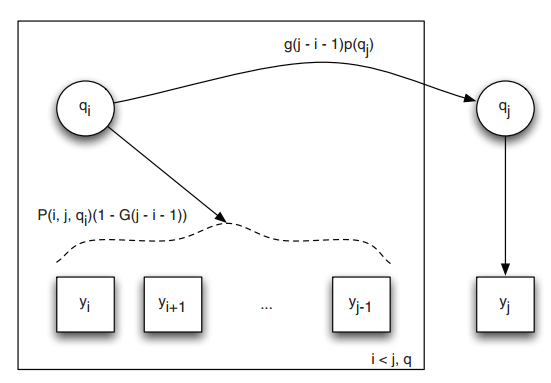}}
\caption{HMM for changepoint detection\cite{konidaris2011robot}. Transitions occur when the model changes.}
\label{fig:hmm}
\end{figure}

This results in a HMM where the hidden state at time $t$ is the model $q_t$ and the observed data is $y_t$ given $x_t$, as shown in Figure \ref{fig:hmm}. The transition from model $q_i$ to $q_j$ occurs with probability
\[ T(q_i, q_j)=g(j-i-1)p(q_j) \]

The emission probability of an observed data segment starting at time $i+1$ and continuing through $j$ using model $q$ is given by
\[ P(y_{i+1}:y_j | q) = P(i,j,q)(1-G(j-i-1)) \]

An online Viterbi algorithm can be used to compute $P_t(j,q)$, the probability of the changepoint previous to time $t$ occurring at time $j$ using model $q$ (i.e., from time $j+1$ to $t$ the data is generated using model $q$) is
\begin{equation} \label{eq: ptjq}
P_t(j,q) = (1-G(t-j-1))P(i,j,q)p(q)P_j^{MAP}
\end{equation}
where $P_j^{MAP}$ is the probability of the MAP changepoint at time $j$,
\[ P_j^{MAP} = \max_{i,q} \frac{P_j(i,q)g(j-i)}{1-G(j-i-1)} \]

Thus at each time $t$, the algorithm computes $P_t(i,q)$ for each model $q$ and changepoint time $j<t$ (using $P_j^{MAP}$) and then computes and stores $P_t^{MAP}$. As $P_t^{MAP}$ being recursively calculated, the MAP changepoint positions and models are stored. When $P_T^{MAP}$ is calculated for the complete observed data sequence, the MAP changepoint positions and models for generating the observed data are identified as a result.

\subsection{Simultaneous Behavior Segmentation and Abstraction Selection}
We can use above changepoint detection algorithm to detect when the abstraction (i.e. $q = \langle O_{rel}, o_{ref} \rangle$) changes and select an appropriate abstraction for each segment. The set of candidate abstractions $Q$ is composed by all possible combinations of $\langle O_{rel}, o_{ref} \rangle$, with the assumption that hand is always relevant, i.e., $h \in O_{rel}$.

Formally, the probability of the behavior segment from time $j+1$ to $t$ conditioned on abstraction $q=\langle O_{rel}, o_{ref} \rangle$ is
\begin{equation}\label{eq:our_pjtq}
P(j,t,q)= p_{ref}(t,o_{ref})\prod_{o_i \in O_{rel}}p_{rel}(j,t,o_i)
\end{equation}
where $n=t-j-1$. And $p_{ref}(t,o_{ref})$ calculates how likely $o_{ref}$ is the correct reference object, based on the distance from the actor(hand in our case) to the object, formally
\[ p_{ref}(t,o_{ref}) = e^{-n\left\lVert loc_t^h - loc_t^{o_{ref}} \right\rVert} \] 
where $loc_t^h, loc_t^{o_{ref}}$ are the 3D locations of the hand and the hypothesized reference object at time $t$. And $p_{rel}(j,t,o_i)$ calculates how likely $o_i$ is a relevant objects, based on whether it is along with hand, formally
\[ p_{rel}(j,t,o_i) =  
  \begin{cases} 
      1 & o_i=h \\
      e^{n(1-\frac{1}{1+e^{-100\overline{d}}})}& \overline{d} \leq d_{thresh} \\
      e^{-\frac{n}{1+e^{-100\overline{d}}}}& else 
   \end{cases}
\]
where $\overline{d}=\frac{\sum_{k=j+1}^{t} \left\lVert loc_k^h - loc_k^{o_i} \right\rVert}{n}$ is the average differences between the displacements of the locations of the hand and object $o_i$ from time $j+1$ to $t$.

For the changepoint detection algorithm to work well, an appropriate model of expected segment length and an appropriate model for fitting the data are required. Similarly to Konidaris et al. \cite{konidaris2011robot}, we assume a geometric distribution for skill lengths with parameter $p$, so that $g(l)=(1-p)^{l-1}p$ and $G(l)=1-(1-p)^l$, and this provides a natural way to set $p$ via $k=1/p$, the expected skill length. 

Using the probability functions defined above, the changepoint detection algorithm can segments the observed behavior into multiple pieces with selected state abstraction, i.e., relevant objects and reference frame.
\section{Policy Learning}\label{sec:policy_learning}
\subsection{Initial RL Problem Formulation}\label{sec:initial_RL}
For $i$th behavior segment, given the selected abstraction  $q = \langle O_{rel}, o_{ref} \rangle$, i.e., relevant objects and reference frame, we can initially formulate the RL problem to be solved as an MDP with components $\langle {\mathcal{S}},{\mathcal{A}},R \rangle$,
\begin{equation}\label{eq:S}
{\mathcal{S}} = \{p^{o_i o_{ref}}|o_i \in O_{rel}\}
\end{equation}
\begin{equation}\label{eq:A}
{\mathcal{A}} = \{a_{pose}(\theta) | \theta \in \Theta_{reachable} \}
\end{equation}
\begin{equation}\label{eq:R}
R: c(s,a) = c_{imm}(a) + c_{ter}(s_T)
\end{equation}
where $p^{o_i o_j}=(x^{ij},y^{ij},z^{ij},qx^{ij},qy^{ij},qz^{ij},qw^{ij})$
 are location and quaternion pose variables of object $o_i$ in object $o_j$ centered reference frame, $a_{pose}(\theta)$ are actions that bring Baxter's end-effector to specific poses $\theta$ following trajectories generated by DMPs, and $\Theta_{reachable}$ are all the end-effector poses that the robot can reach to.
 
Since the robot doesn't know whether the private information such hand holding force is important or not in advance, thus by default it assumes the private information is not important at first. If later there is enough evidence that the RL problem needs to be reformulated, then the {\em private action} $a_{gripper}$, which controls the holding force $\lambda$ of the robot gripper($\lambda=0$ for opening gripper, $\lambda>0$ for closing gripper with force $\lambda$),is added into the action space $\mathcal{A}$ in the reformulation stage, as elaborated later in section \ref{sec:reformulation}.
 
The cost function $c_{total}$ is composed by immediate cost function $c_{imm}(a)$ and terminal cost function $c_{ter}(s_T)$,
\begin{equation}\label{eq:cimm}
c_{imm}(a)=w_{imm}^T \ddot{a}
\end{equation}
\begin{equation}\label{eq:cter}
c_{ter}(s_T) = w_{ter}^T (s_T - s_T^{obs})
\end{equation}

where $c_{imm}(a)$ penalizes large accelerations, and $c_{ter}(s_T)$ penalizes large differences between the terminal state $s_T$ at the end of the executions of actions, and the terminal state $s_T^{obs}$ observed at the end of the behavior to be imitated. Note that we also penalize gripper opening/closing action with cost at 5. Thus the total cost of a trial is calculated as the sum of the immediate cost and the terminal cost.

\subsection{RL with DMP}
We represent the movements of the end-effector of a robot through a set of
Dynamic Movement Primitives(DMPs), as introduced by Schaal \cite{schaal2006dynamic} and Ijspeert\cite{ijspeert2013dynamical}. Thus the actions $a_{pose}$ are executed through position control following the pose trajectories generated by a set of DMPs. 

Just a quick recap on DMP, there are discrete DMPs(i.e., point to point) and rhythmic DMPs(periodic), and here we are focused on discrete DMPs. The discrete DMP is formulated as a point attractor system modulated by nonlinear terms such that it achieves a desired attractor behavior, which is appropriate for object manipulation tasks.

We used a modified version of DMP formulation as proposed in \cite{hoffmann2009biologically},
\[ \tau \dot{z} = \alpha_z(\beta_z(g-y)-z) - \alpha_z \beta_z (g-y_0)x + \alpha_z \beta_z f \]
\[ \tau \dot{y} = z \]
these equations represent the dynamical system, where $\tau$ is a time constant, $\alpha_z$ is damping constant, and $\alpha_z \beta_z$ is spring constant. $y, \dot{y}, \ddot{y}$ can be interpreted as desired position, velocity, and acceleration for a control system, and a controller would convert these variables into motor commands.

The forcing term $f$ is how we can adjust the landscape of the original point attractor represented by the damped spring model. And $f$ is formulated as scaled linear combination of basis function $\psi_i(x)$ weighted by $w_i$
\[ f(x) =\frac{\sum_{i=1}^N \psi_i(x)w_i}{\sum_{i=1}^N \psi_i(x)} x \]
\[ \psi_i(x) =exp(-\frac{1}{2\sigma_i^2}(x-c_i)^2) \]
where the weight vector is $\mathbf{w}=[w_1,w_2,\cdots,w_N]^T$, $\sigma_i$ and $c_i$ are constants that determine, respectively, the width and centers of the basis functions and $y_0$ is the initial state, i.e., $y_0= y(t = 0)$, and
\begin{equation}\label{eq:canonical}
\tau \dot{x} = -\alpha_x x
\end{equation}
this equation is called the canonical system because it models the generic behavior of the dynamical system, a point attractor in our case and a limit cycle for rhythmic behavior. We can regard $x$ as a phase variable that replaces the explicit timing, and $x$ can start at some arbitrary initial state (typically 1).

As we can see, in the context of RL, assuming the basis functions $\psi_i(x)$ and all the other constants in DMPs are given, the action is parametrized by the weight vector $\mathbf{W}$, which is composed by concatenating all the weight vectors $\mathbf{w}$ for all the DMPs, e.g., we are using 7 DMPs for 7 DOFs of the robot end-effector. Thus essentially policy search methods can be applied to learn the optimal policy parametrized by the weight vector $\mathbf{W}$.

\begin{figure}[h!]
\centering
\includegraphics[width=1.0\textwidth]{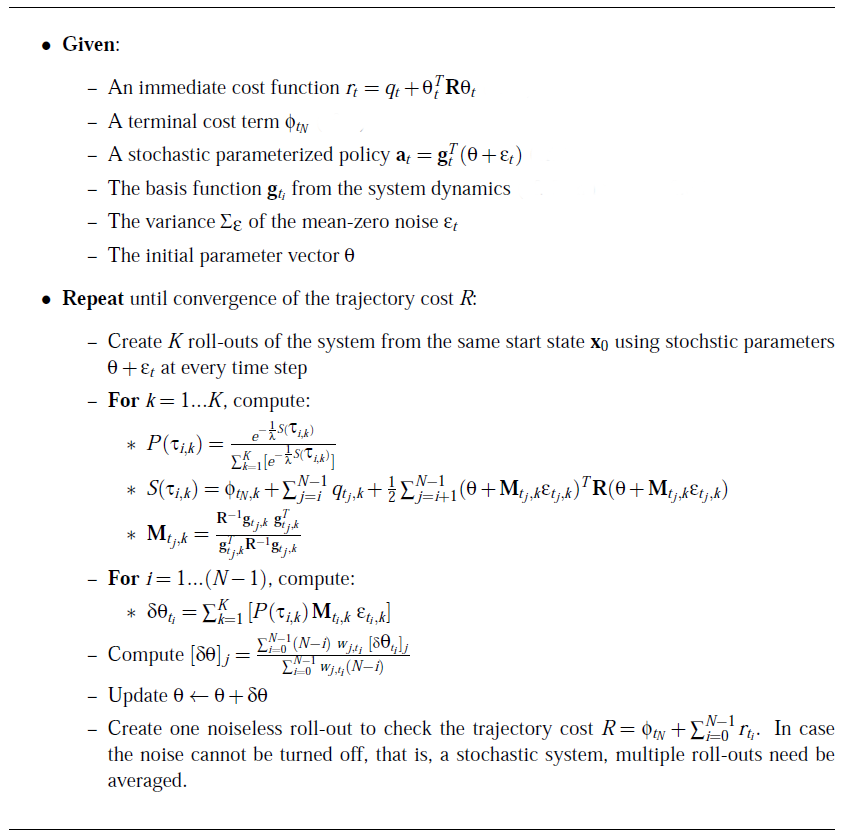}
\caption{Pseudocode of PI$^2$ algorithm \cite{theodorou2010generalized}. In detail, $q_t=q(\mathbf{x_t})$ is an arbitrary state-dependent cost function, where $\mathbf{x_t}$ denotes the state of the system. $\mathbf{R}$ is a positive semi-definite weight matrix of the quadratic control cost, $\boldsymbol{\epsilon_t}$ is the Gaussian noise vector, $N$ is the length of one roll-out, and $\uptau_i=(\mathbf{x_{t_i}}, \cdots, \mathbf{x_{t_N}})$ is a sample path starting at state $\mathbf{x_{t_i}}$.}
\label{fig:pi2_alg}
\end{figure}

Theodorou et al. proposed Policy Improvement with Path Integrals(PI$^2$)\cite{theodorou2010reinforcement}\cite{theodorou2010generalized}
that can improve policy parametrized as a Dynamic Movement Primitive(DMP)\cite{ijspeert2013dynamical}. PI$^2$ transforms policy improvements into an approximation problem of a path integral. The pseudocode of the algorithm is as shown in Figure \ref{fig:pi2_alg}, they used some different notations for DMP elements in their context, to clarify, the policy parameters $\boldsymbol{\theta}$ in their algorithm is equivalently the weight vector $\mathbf{w}$ used here in one DMP formulation, and action $a$ in their algorithm corresponds to $f$ used here in DMP formulation. For more details, please refer to \cite{theodorou2010reinforcement}\cite{theodorou2010generalized}.

As a result of PI$^2$, for $i$th initially formulated RL problem, we can learn the optimal policy $\pi_i$ as parametrized by the learned weight vector $\mathbf{w}$. We store the learned policy $\pi_i$ together with initiation set ${\mathcal{I}}_i$ and termination condition $\beta_i$ as a learned option $o_i: \langle {\mathcal{I}}_i, \pi_i, \beta_i \rangle$,
\[
 \beta_i(s) =
  \begin{cases} 
      \hfill 1    \hfill & \text{if} c_{ter}(s)<C  \\
      \hfill 0 \hfill & \text{else} \\
  \end{cases}
\]
where $C$ is a threshold on the total cost. Initiation set ${\mathcal{I}}_i$ is composed by the states that the robot encountered that can lead to $\beta_i(s_T)=1$ following $\pi_i$. We used $C=20$ in our experiment.

\section{RL Problem Reformulation}\label{sec:reformulation}
As a criteria evaluating whether the $i$th segmented behavior is successfully imitated, we use $P(\beta_i(s_T)=1)|\pi_i)$, i.e., the probability of successfully reproduced the observed behavior following policy $\pi_i$, to decide whether initially formulated RL problems need to be reformulated. 

If $P(\beta_i(s_T)=1)|\pi_i)<\alpha$, where $\alpha$ is a pre-defined threshold for determining whether the imitation is successful, then that indicates that 1)there are large accelerations during the action execution, and 2)at the end of executions of actions following learned $\pi_i$, the terminal state $s_T$ is often far from the observed terminal state $s_T^{obs}$ in the observed behavior to be imitated, then the initially proposed RL problem needs to be reformulated to capture missing aspects of the behavior such that a better policy can be learned, and the RL problem is reformulated until $P(\beta_i(s_T)=1)|\pi_i)>\alpha$ for all segmented behaviors. 

Our RL reformulation strategy is described in \ref{alg:reformulation}. To search for the correct RL problem formulation for the $i$th segmented behavior, we first start with an initial RL problem formulation based on the selected abstraction, as explained earlier in section \ref{sec:initial_RL}. If the policy learned for the initially formulated RL problem is not good enough to reproduce the observed reliably, then we propose to reformulate the RL problem for the $i$th segmented behavior. We first look at the likelihood of the observed data in the $i$th segment given all the possible abstractions, calculated as in equation \ref{eq: ptjq}, then sort all the abstractions that give likelihood higher than a pre-defined threshold $\alpha_{model}$ and store them in $Q$. The start to reformulate the RL problem using ranked abstractions in $Q$, if the first abstraction doesn't work out, then try reformulating the RL problem by including gripper action $a_{gripper}$ in, which controls the holding force of the grippers; if that doesn't work out, then try the second abstraction, and so on, until the reformulation works out. If unfortunately, all the abstractions in $Q$ have been tested and failed, then the failure of imitation could due to inappropriate RL problem formulation for earlier behavior segments, thus proceed to reformulate the $i-1$th segmented behavior.

\begin{algorithm}[h!]
\caption{RL Reformulation}\label{alg:reformulation}
\begin{algorithmic}
\Procedure{Searching Correct RL Problem Formulation}{}
\State Given $i$th segmented behavior start from time $t_i+1$ to $t_{i+1}$ 
\State Proposed an initial RL formulation $\langle {\mathcal{S}}_i,{\mathcal{A}}_i,R_i \rangle$ 
\State Learn the optimal policy for initially formulated RL problem
\While {$P(\beta_i(\mathbf{s}_T)=1)|\pi_i)<\alpha$}
	\State \textbf{Reformulate($i$)}
	\State learn the optimal policy for reformulated RL problem
\EndWhile
\State Store the learned option  $o_i: \langle {\mathcal{I}}_i, \pi_i, \beta_i \rangle$
\EndProcedure

\\

\Function{\textbf{Reformulate($i$)} \Comment{reformulate RL problem for the $i$th segment}}{}
\State \textbf{global} $reform_{curr}$ (default \textit{false})
\State \textbf{global} $reform_{prev}$ (default \textit{false})
\If {$reform_{prev}$}
	\State \textbf{Reformulate($i-1$)}
	\State $reform_{prev}=false$
\EndIf
\If {$!reform_{curr}$}
	\State calculate $P_{t_{i+1}}(t_i,q) \forall q$ based on equation \ref{eq: ptjq}
	\State $Q = \{ q | \frac{P_{t_{i+1}}(t_i,q)}{\max_q P_{t_{i+1}}(t_i,q)} > \alpha_{model} \}$ 
	\State sort $q \in Q$ based on $P_{t_{i+1}}(t_i,q)$
	\State $A_i\prime = \{a_{pose}, a_{gripper}\}$
	\State $M = \{\langle {\mathcal{S}}_i,{\mathcal{A}}_i,R_i \rangle, \langle {\mathcal{S}}_i, {\mathcal{A}}_i\prime, R_i \rangle | q \in Q \}$ based on equations \ref{eq:S}, \ref{eq:A}, \ref{eq:R}	
	\State $reform_{curr}=true$
\EndIf
\If{$M \neq \emptyset$}{}
	\State $\langle {\mathcal{S}},{\mathcal{A}},R \rangle = pop(M)$
	\State Reformulate the $i$th RL problem to $\langle {\mathcal{S}},{\mathcal{A}},R \rangle$
\Else
	\State $reform_{curr}=false$
	\State $reform_{prev}=true$
\EndIf
\EndFunction
\end{algorithmic}
\end{algorithm}
\section{Experiment}\label{sec:experiment}
In our experiment, we show that by observing single human demonstration on stacking two blocks together, as shown in Figure \ref{fig:demo}, the robot can automatically segment the observed behavior into pieces, and formulate correct RL problems by recovering from initial failure of imitation due to inappropriate initial RL problem formulation, and in the end, learn good policies such that the observed behavior is successfully reproduced.

\begin{figure*}[h!]
\centering
\subfigure[]{
\includegraphics[height=0.37\textwidth, width=0.48\textwidth]{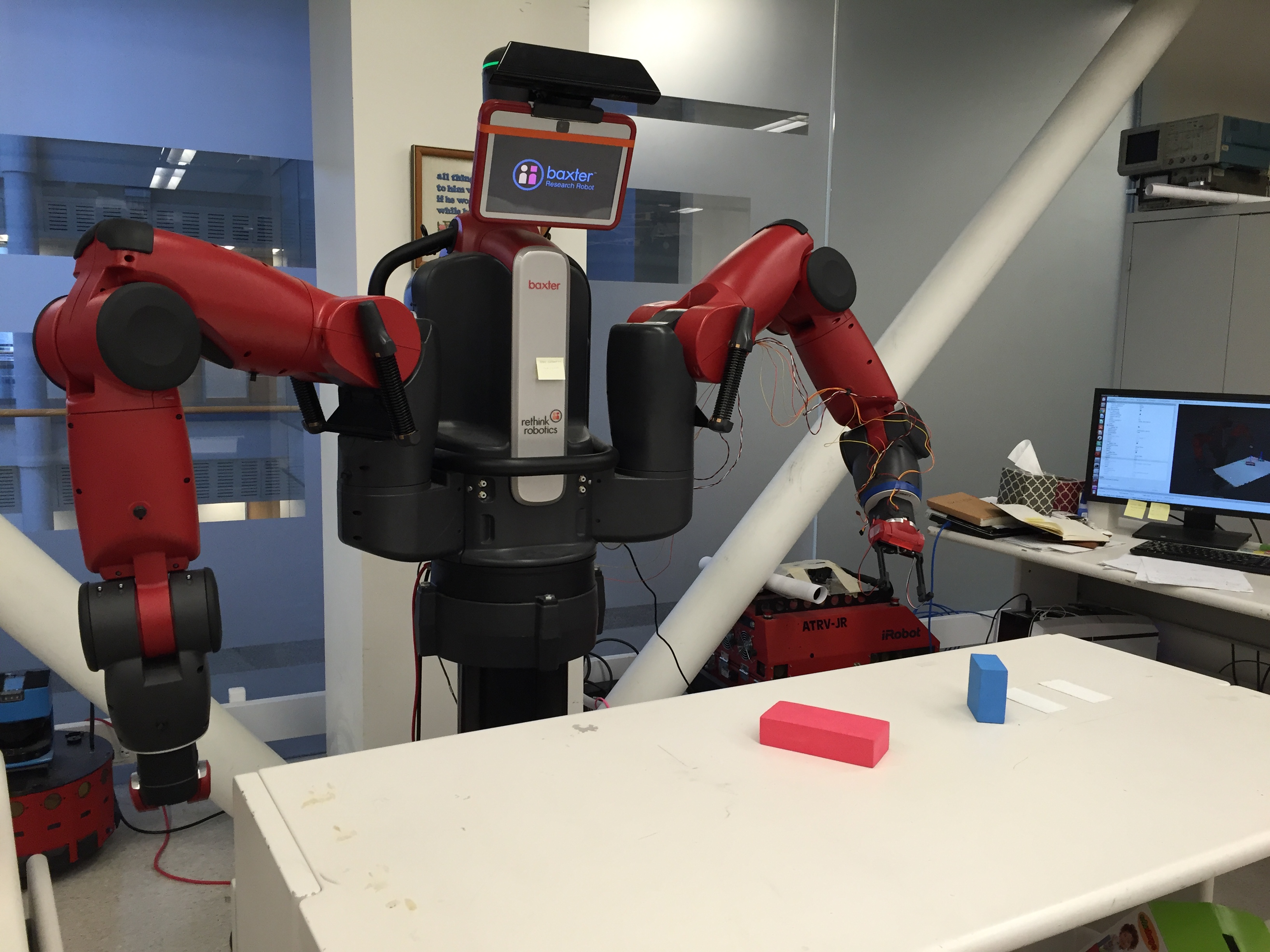} 
}
\subfigure[]{
\includegraphics[height=0.37\textwidth, width=0.48\textwidth]{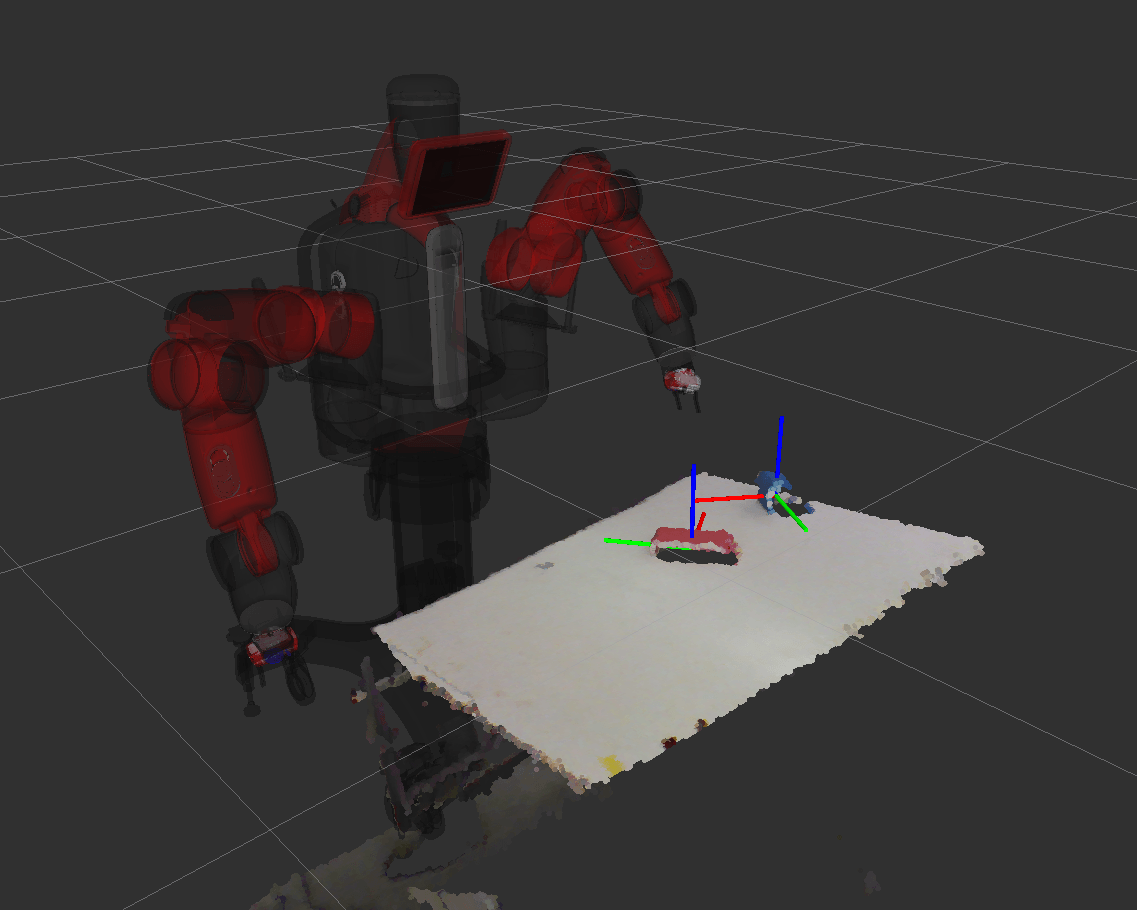} 
}
\caption{(a)Experiment setup. (b)ROS Rviz view of point cloud messages from Kinect.\label{fig:setup}}
\end{figure*}

Our experiment was carried out on our physical robot, Baxter Research Robot \cite{baxter}, we used a Kinect sensor mounted on the head of Baxter, as shown in Fig. \ref{fig:setup}(a). The Kinect sensor records the RGB-D frames of the human demonstration and robot imitation trails. One example of the point cloud view from the Kinect sensor is shown in Fig. \ref{fig:setup}(b) using ROS Rviz\footnote{ROS package: \url{http://wiki.ros.org/rviz}}. The Kinect sensor is calibrated with the robot such that the transformation between the kinect camera coordinate frame and the robot coordinate frame is known.

Next, we'll discuss how the human behavior observed through Kinect is processed, and how Baxter control its arms for action execution, and then the experiment results and discussion.

\subsection{Perception}
To extract the sequence of states \ref{eq:observation} from the RGB-D video, we need to 1) locate the objects and hand at the 1st frame, and 2) track the objects and hand, such that we can record the 3D poses of objects and hand. The tracking results are as shown in Figure \ref{fig:demo}, and the detailed visual processing is as elaborated in Appendix \ref{app:vision}.

\begin{figure}
\centering
\begin{tabular}{@{}ccc@{}}
  \begin{tabular}{@{}c@{}}
  \includegraphics[height=0.25\textwidth, width=0.31\textwidth]{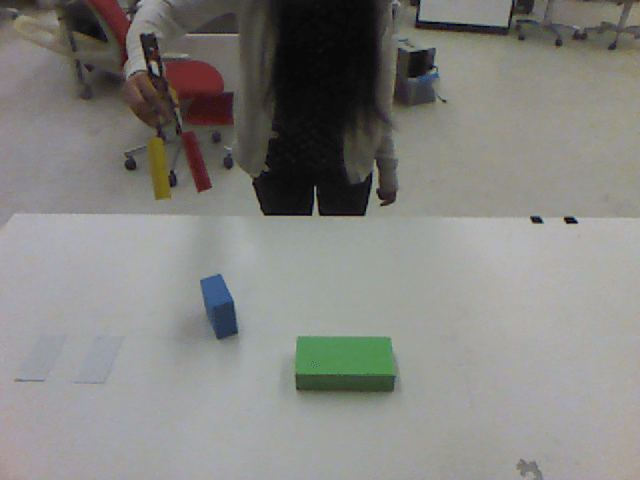}\\
  \includegraphics[height=0.25\textwidth, width=0.31\textwidth]{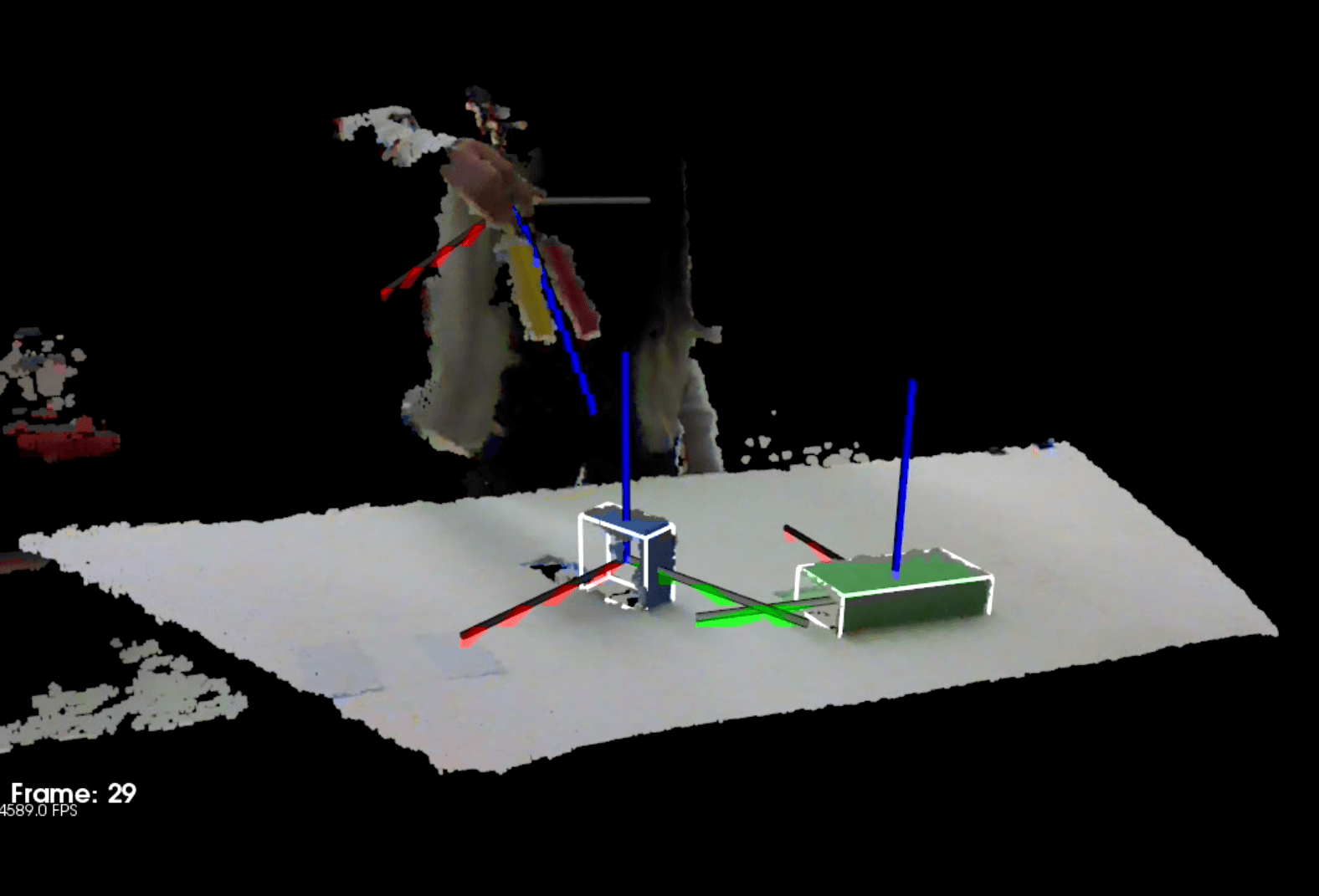}
  \end{tabular}
  &
  \begin{tabular}{@{}c@{}}
  \includegraphics[height=0.25\textwidth, width=0.31\textwidth]{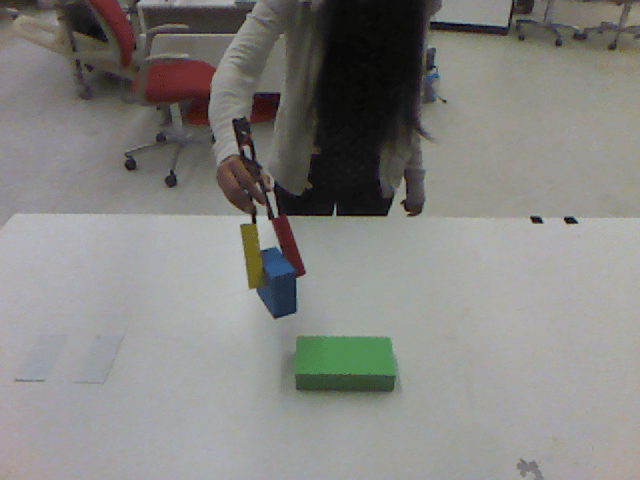}\\
  \includegraphics[height=0.25\textwidth, width=0.31\textwidth]{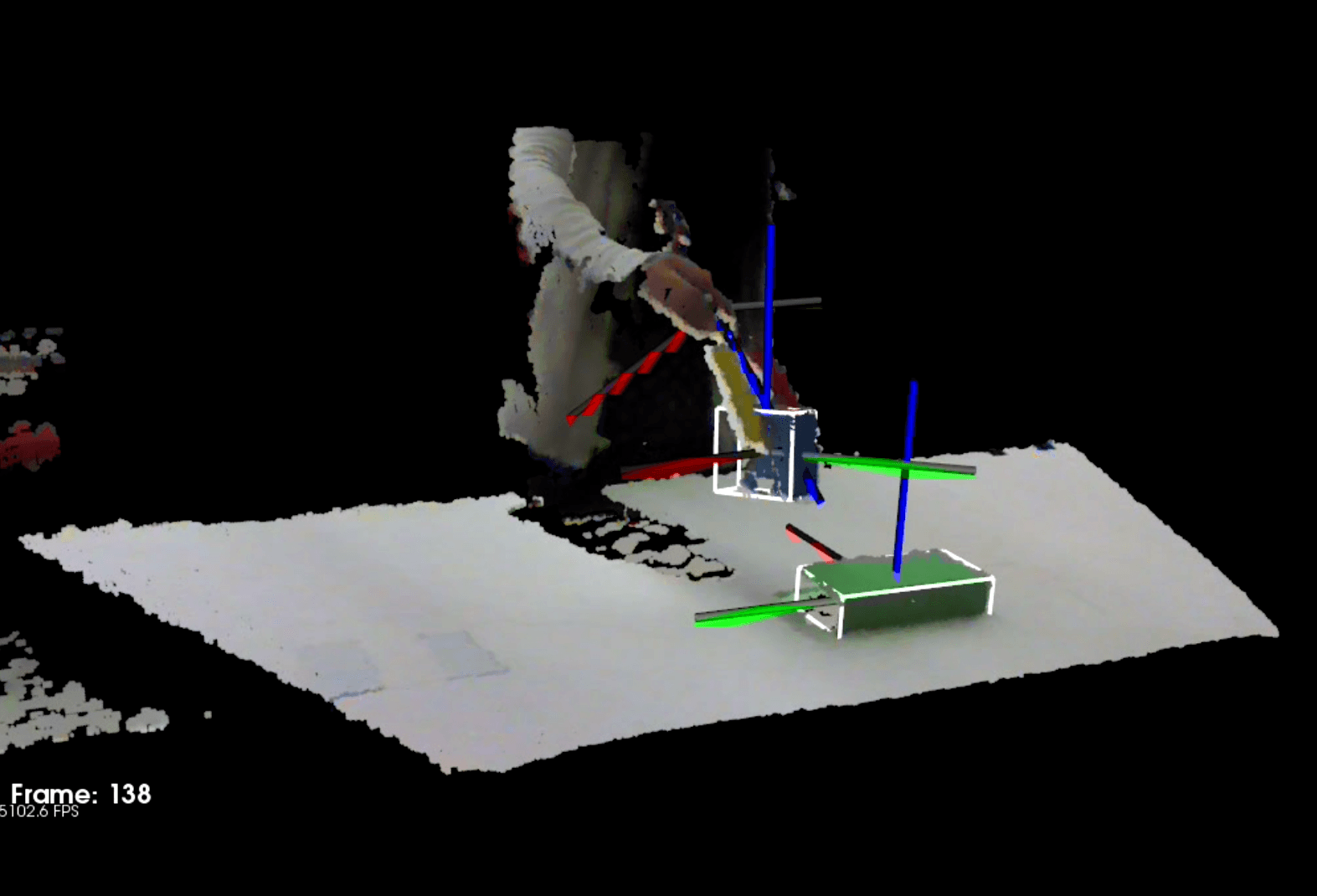}
  \end{tabular}
  &
  \begin{tabular}{@{}c@{}}
  \includegraphics[height=0.25\textwidth, width=0.31\textwidth]{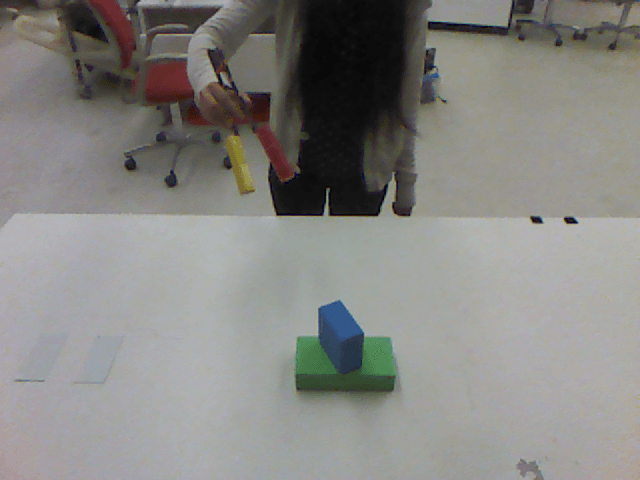}\\
  \includegraphics[height=0.25\textwidth, width=0.31\textwidth]{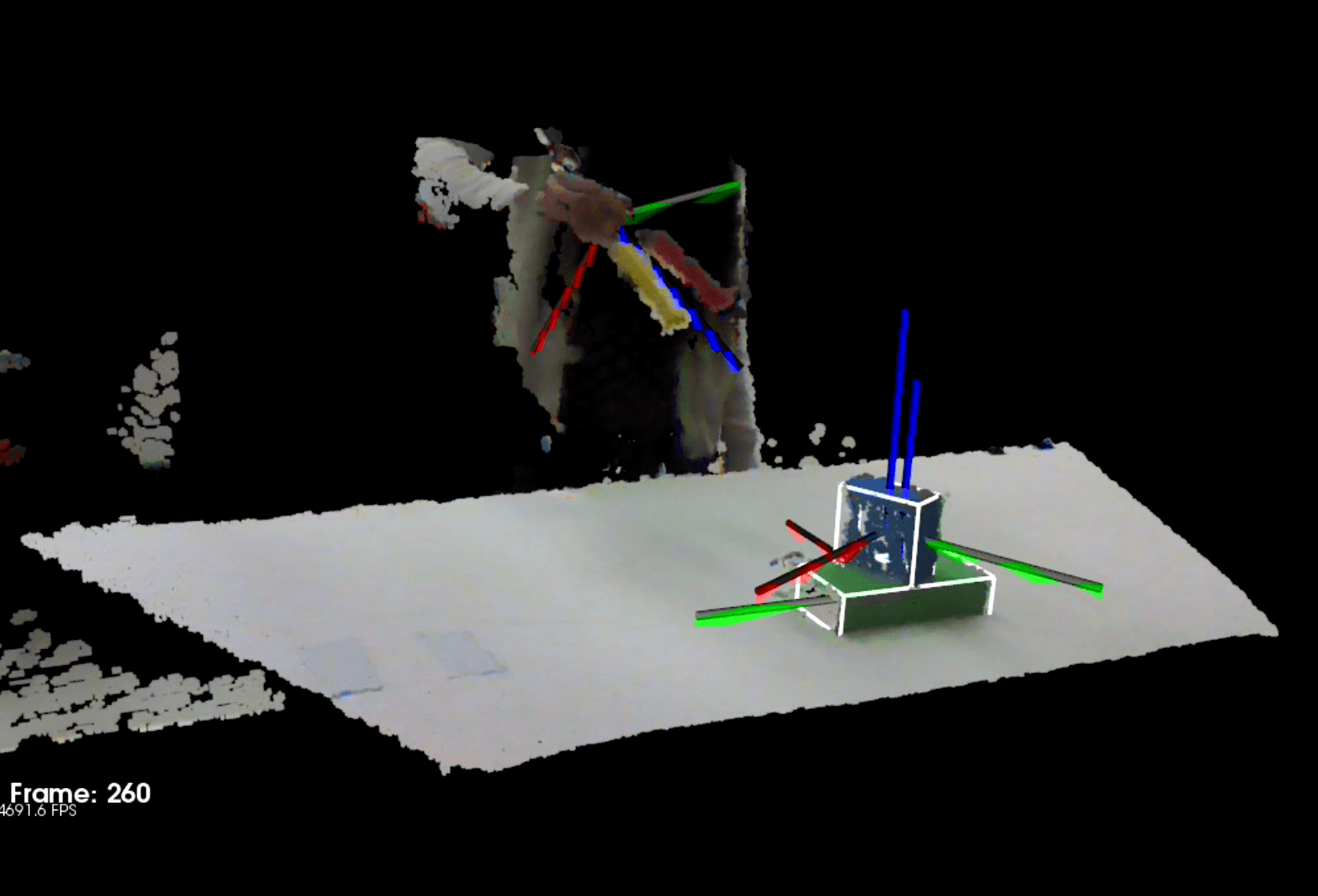}
  \end{tabular}
\end{tabular}

\caption{\textit{Upper row}: observed behavior; \textit{Lower row}: tracked objects and hand. From left to right, the actions are 1)reaching to grasp blue block, 2)carrying blue block and stack it onto green block, 3)retrieve hand from blocks. The full-length video is available at \url{https://youtu.be/mogAk8ndNek}, and the tracking result is available at \url{https://youtu.be/-hLooluR9zA}.}\label{fig:demo}

\end{figure}

\subsection{Control}
In our experiment, we use 7 DMPs to control the 7 DOFs of Baxter end-effector, i.e., the 3D position and the orientation expressed in quaternion, and an extra DMP for robot gripper holding force. We share one canonical system \ref{eq:canonical} among all DMPs so that the canonical system provides the temporal coupling between all DOFs, while each dynamical system associated with each DOF presents the desired attractor behavior.

For the control of the Baxter end-effector, we use 7 DMPs to generate a 3D pose trajectory based on the policy parameters being learned, then inverse kinematics is first applied to get the corresponding arm joints angle trajectory, then we can directly send the joints angle commands to Baxter and the internal controllers \footnote{\url{http://sdk.rethinkrobotics.com/wiki/Arm_Control_Modes}} will take care of following the joints angle trajectory.

Based on the comparison between the original DMP formulation \cite{schaal2006dynamic} and the biologically-inspired DMP formulation \cite{hoffmann2009biologically} as explained in Appendix \ref{app:dmp_vs}, we use the latter in our work.

For more details on DMP, please refer to the technical report \cite{zhenz_dmpmath_2015}.

\subsection{RL Methods}\label{sec:initial_policy}
As explained earlier in Sec \ref{sec:policy_learning}, we applied PI$^2$ algorithm to learn our policy parameters. The reason why we choose PI$^2$ over another promising method PoWER \cite{kober2009policy}\cite{kober2009learning} is as explained in Appendix \ref{app:power_vs_pi2}.

Specifically, the policy parameters get updated for the first time after the first 10 trials, and after that, the policy parameters get updated every other 5 trials. The maximum number of updates is 5. And the policy is considered to be converged when the difference between the total costs $R(s,a)$(equation \ref{eq:R}) of consecutive updates is smaller than 3.

The policy parameters are initialized by learning the weight vectors of the 7 DMPs such that those DMPs can generate trajectories that resemble the shape of the observed trajectory. For the details on getting the initial policy parameters, please refer to Sec 2.2 in technical report \cite{zhenz_dmpmath_2015}.

\subsection{Results}
\subsubsection{Behavior Segmentation and Abstraction Selection} \label{sec: vs_konidaris}
\begin{figure}[h!]
\centering
\centerline{\includegraphics[width=1.2\textwidth, height=0.4\textheight]{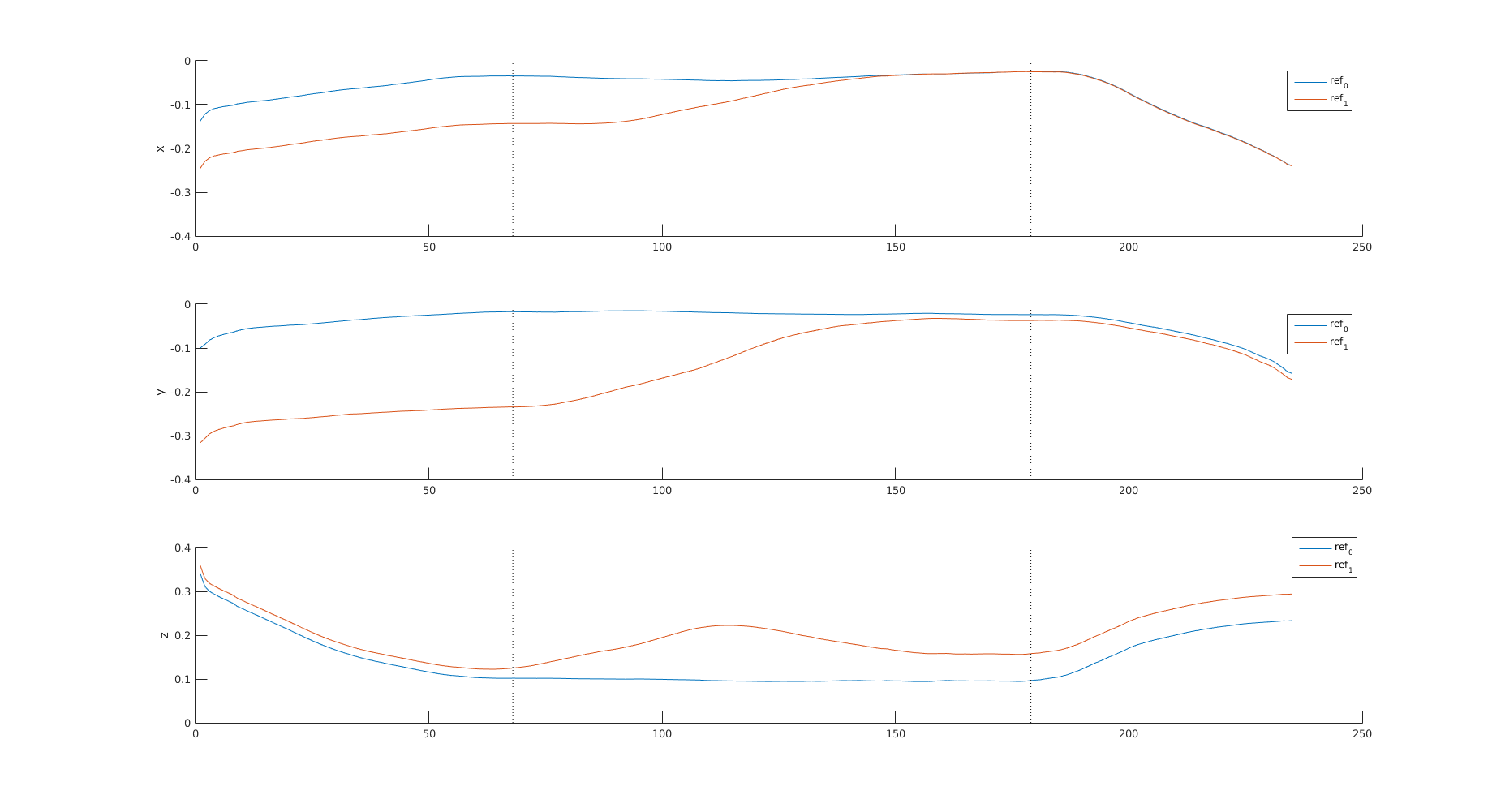}}
\caption{Original hand location trajectories in \textit{block0}(blue) and \textit{block1}(red) reference frame. Our algorithm selects the abstraction in the form of $\langle O_{rel}, o_{ref} \rangle$ in sequence: $\langle \{h\},block0 \rangle, \langle \{h, block1\},block0 \rangle, \langle \{h\},block0 \rangle$.}
\label{fig:test10_ours}
\end{figure}

Given the blocks and hand poses extracted from the single human demonstration on stacking two blocks, we applied our method as explained in Sec \ref{sec:segmentation} to the pose trajectories, and the result is as shown in Figure \ref{fig:test10_ours}. Note that the position of the hand is plotted in the reference frame of the blue block and green block, such that it is easier to interpret the meaning of the trajectory for the reader.

Our method segment the observed behavior into three pieces: 1) the 1st piece corresponds to hand reaching towards the blue block and end with a grasp pose; 2) the 2nd piece corresponds to hand stacking the blue block onto the green block; 3) the 3rd piece corresponds to hand retrieving from those blocks. For these three pieces, our method selects the abstraction in the form of $\langle O_{rel}, o_{ref} \rangle$ correctly, 1) the 1st piece is associated with $O_{rel}$ being the hand, $o_{ref}$ being the blue block; 2) the 2nd piece is associated with $O_{rel}$ being the hand and the blue block, $o_{ref}$ being the green block; 3) the 3rd piece is associated with $O_{rel}$ being the hand, $o_{ref}$ being the blue block.

\begin{figure}
\centering
\centerline{\includegraphics[width=1.2\textwidth, height=0.4\textheight]{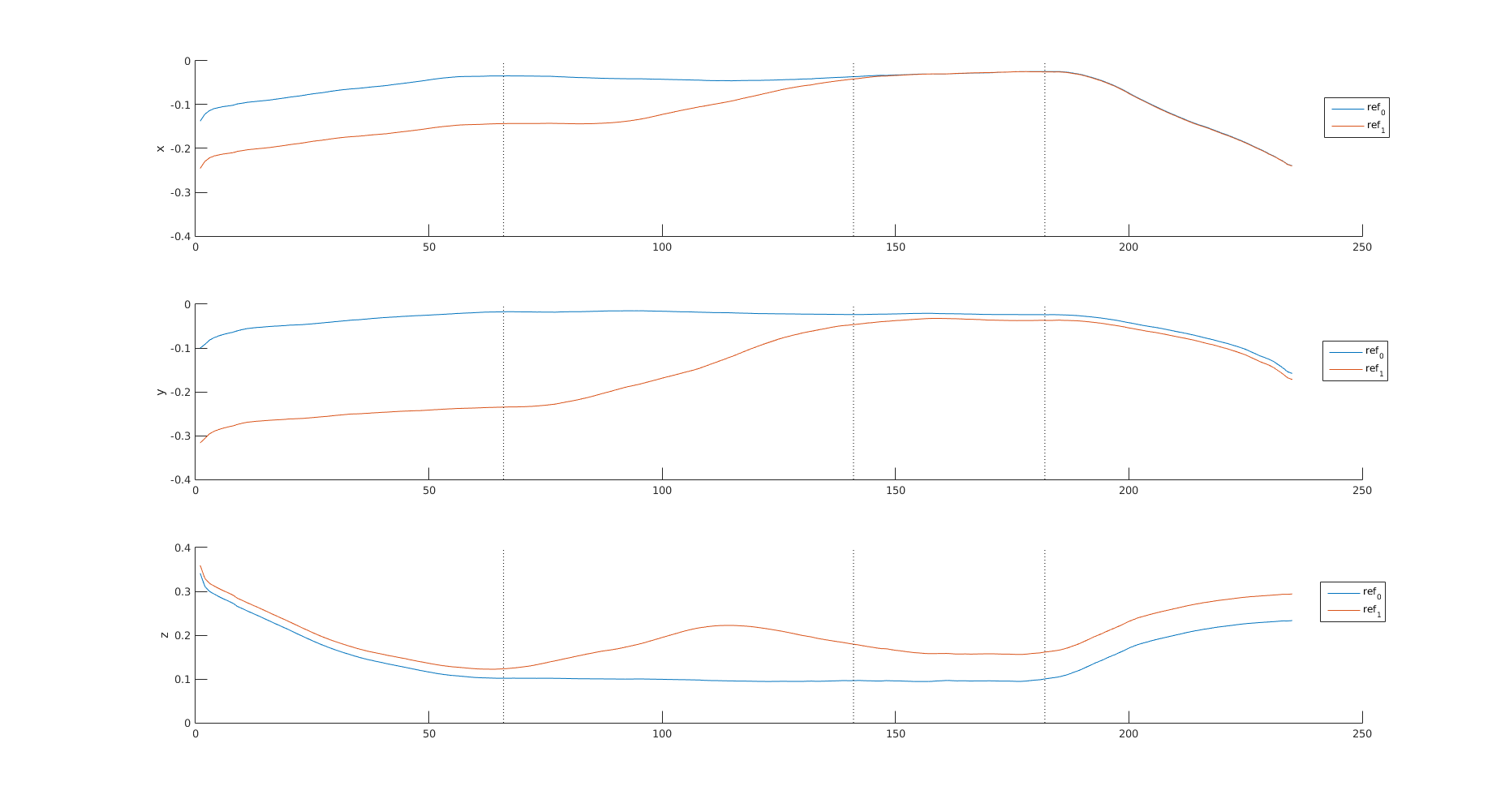}}
\caption{Original hand location trajectories in \textit{block0}(blue) and \textit{block1}(red) reference frame. Konidaris's algorithm selects the abstraction in the form of $\langle O_{rel}, o_{ref} \rangle$ in sequence: $\langle \{h\},block1 \rangle$, $\langle \{h\},block1 \rangle$, $\langle \{h\},block1 \rangle$, $\langle \{h\},block1 \rangle$.}
\label{fig:test10_konidaris}
\end{figure}

While Konidaris's method \cite{konidaris2011robot}, when applied to the same pose trajectories, returned wrong results as shown in Figure \ref{fig:test10_konidaris}. Their method is based on the intuition that the observed behavior should be segmented into pieces such that each piece has a different underlying policy, thus they approximate the value function for each state encountered during a temporal segment of the observed behavior, and divide the observed behavior into pieces when a single value function cannot approximate the observed rewards well enough. 

Their method segments the observed behavior into 4 pieces, with the first and the last piece being similar to our result. But their method oversegments the "stacking" trajectory into two pieces, where the 1st piece is mostly the "stacking" trajectory, and the 2nd piece is hand being steady for a while due to slow motion of the hand in the observed human demonstration for that particular time period, this is due to that these two pieces seem to have different underlying policy, one being "moving the hand", the other being "keeping the head steady". In addition, their method selects the wrong object-centered reference frame for the first segment, where the hand is reaching towards the blue block. An analysis of the failure modes of their method is provided in this technical report \cite{zhenz_debug_konidaris_2015}.

\subsubsection{Imitation Results with Initial RL Problem Formulation}
Given the segmented behavior and the state abstraction selected for each behavior segment, we formulate an initial RL problem for each segment, following the rules as explained in equations \ref{eq:S}, \ref{eq:A}, \ref{eq:R}.


Starting from the 1st behavior segment, the initial RL problem is formualted as: ${\mathcal{S}}=p^{h o_{blue}}$, $\mathcal{A}=\{a_{pose}(\theta) | \theta \in \Theta_{reachable}\}$, $R: c(s,a) = c_{imm}(a) + c_{ter}(p^{h o_{blue}}_T)$. Baxter first estimates the initial policy based on the observation as explained in Sec \ref{sec:initial_policy}, and then PI$^2$ is applied to learn the updates of the policy parameters such that the total cost $R(s,a)$(equation \ref{eq:R}) can be decreased. Once the policy is converged, we measure how likely the robot can reproduce the observed behavior, i.e., $P(\beta_1(s_T)=1)|\pi_1)$. If $P(\beta_1(s_T)=1)|\pi_1)>\alpha$, we claim that the robot has successfully reproduced the 1st behavior segment, otherwise the robot needs to reformulate the RL problem following the algorithm \ref{alg:reformulation}.

\begin{figure*}[h!]
\centering
\subfigure[]{\label{fig:initial_cost}
\includegraphics[height=0.37\textwidth, width=0.48\textwidth]{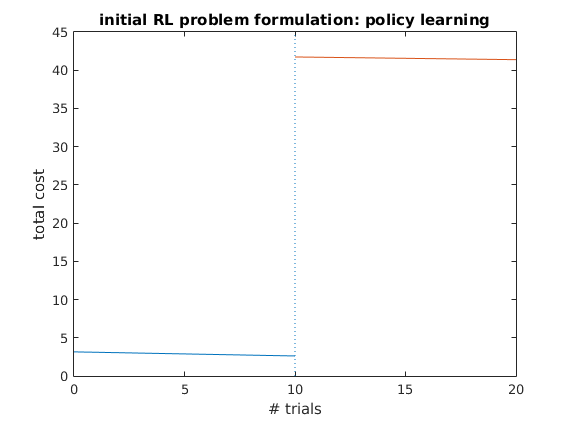} 
}
\subfigure[]{\label{fig:reformulated_cost}
\includegraphics[height=0.37\textwidth, width=0.48\textwidth]{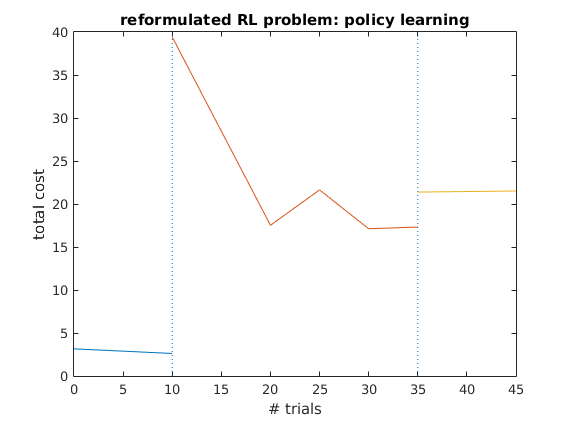} 
}
\caption{Total cost accumulated over number of trials on learning different behavior segments. \textit{Blue}: 1st behavior segment - end-effector reaches towards the blue block with a grasp pose; \textit{Red}: 2nd behavior segment - carry the blue block to stack it onto the green block; \textit{Red}: 3rd behavior segment - retrieve the end-effector from the blocks.
(a) Learning curve based on initial RL problem formulation. \textit{Blue}: For the learning of the 1st behavior segment, policy $\pi_1$ is converged after 10 trials, and $P(\beta_1(s_T)=1)|\pi_1)=1$ thus no need for RL problem reformulation; \textit{Red}: for the learning of the 2nd behavior segment, policy $\pi_2$ is converged after 10 trials, while $P(\beta_2(s_T)=1)|\pi_2)=0$, thus the robot needs to reformulate the initially proposed RL problem until it can successfully reproduce the 2nd behavior segment, before it starts to learn to reproduce the 3rd behaviro segment.
(b) Learning curve based on reformulated RL problem. \textit{Blue}: The learning for the 1st behavior segment is completed with the initial RL problem formulation as in (a), thus this is the same cost trace be presented in (a). \textit{Red}: For the learning of the 2nd behavior segment given the reformulated RL problem, the policy is converged after 25 trials, the total cost of the 2nd behavior segment gets down to around 15, and the accumulated cost across the 1st and the 2nd behavior segment gets down to around 17, which is much less than the cost in (a). And $P(\beta_2(s_T)=1)|\pi_2)=1$, thus no need for further RL problem reformulation. \textit{Orange}: For the learning of the 3rd behavior segment, the policy is converged after 10 trials, and $P(\beta_2(s_T)=1)|\pi_2)=1$, thus there is no need for RL problem reformulation.}
\end{figure*}

We use $\alpha=0.8$ in our experiment, and Baxter successfully learned to reach to the blue block with a grasp pose, as show in video \url{https://www.dropbox.com/s/lv685yx927nus4w/wrong_a_rviz_speedup.mp4?dl=0}. And the learning curve for the 1st behavior segment is as shown by the blue line in Figure \ref{fig:initial_cost}.

Then Baxter carried on to learn the policy for the 2nd behavior segment, following the same learning process it did for the 1st behavior segment as explained above. The initial RL problem is formulated as: ${\mathcal{S}}=\{p^{h o_{green}}, p^{o_{blue} o_{green}}\}$, $\mathcal{A}=\{a_{pose}(\theta) | \theta \in \Theta_{reachable}\}$, $R: c(s,a) = c_{imm}(a) + c_{ter}(p^{h o_{green}}_T, p^{o_{blue} o_{green}}_T)$. When the policy converged, Baxter failed to reproduce the observed behavior with $P(\beta_2(\mathbf{s}_T)=1)|\pi_2)=0$. The learning curve is as shown by the red line in Figure \ref{fig:initial_cost}.

The failure is due to the fact that the private action $a_{gripper}$ is not included in the action sapce $\mathcal{A}$ in the initial RL problem formulation. To carry the blue block around, Baxter needs to close its gripper first to grasp the blue block. Thus for the 2nd behavior segment, Baxter failed to reproduce the observed behavior, as shown in video \url{https://www.dropbox.com/s/lv685yx927nus4w/wrong_a_rviz_speedup.mp4?dl=0}.

\subsubsection{Imitation Results with Reformulated RL Problem}
Following the reformulation strategy \ref{alg:reformulation} as discussed earlier, Baxter reformulated the RL problem for the 2nd behavior segment by adding the private action $a_{gripper}$ into the action space, i.e., $\mathcal{A}=\{a_{pose}(\theta), a_{gripper} | \theta \in \Theta_{reachable}\}$. Since the private action is not observed, we cannot initialize the policy parameters for the DMP that generates the value trajectory of the gripper holding force as we did for the DMPs that generate the pose trajectory of the end-effector, thus we initialize the policy parameters associated with $a_{gripper}$ as 0. And we initialize the policy parameters associated with $a_{pose}$ as the ones learned in the initial RL problem formulation. Baxter ended up learning a good policy to reproduce the 2nd behavior segment, with $P(\beta_2(s_T)=1)|\pi_2)=1$, and the learning curve is as shown by the red line in Figure \ref{fig:reformulated_cost}.

Then Baxter carried on to learn to reproduce the 3rd behavior segment, which is to retrieve its end-effector from the blocks. The RL problem for the 3rd behavior segment is initially formulated as: ${\mathcal{S}}=p^{h o_{blue}}$, $\mathcal{A}=\{a_{pose}(\theta) | \theta \in \Theta_{reachable}\}$, $R: c(s,a) = c_{imm}(a) + c_{ter}(p^{h o_{blue}}_T)$. When the policy converged, Baxter succeeded to reproduce the observed behavior with $P(\beta_3(\mathbf{s}_T)=1)|\pi_3)=1$. The learning curve is as shown by the orange line in Figure \ref{fig:reformulated_cost}.

Note that the final policy $\pi_2$ for the 2nd behavior segment has the robot close its gripper around the beginning, before the arm starts to move, such that the robot can grasp the blue block. One interesting thing is that $\pi_2$ happens to have the robot open its gripper around the end, thus releasing the blue block onto the green block, so when learning to reproduce the 3rd behavior segment, the robot doesn't need to worry about the private action, and the initial RL problem formulation for the 3rd behavior segment is good enough.

Thus in the end, based on simultaneous behavior segmentation and abstraction selection, and automatic RL problem reformulation when needed, Baxter was able to reproduce the observed stacking behavior successfully, as shown in video \url{https://www.dropbox.com/s/oc0v93k7e8craty/stack_rviz_speedup.mp4?dl=0}.

\section{Discussions}
We showed that our method can segment an observed behavior into appropriate pieces and select the correct state abstraction for each piece, while Konidaris's method \cite{konidaris2011robot} failed to segment the observed behavior into reasonable pieces and select the correct state abstraction for each piece. We argue that our method is more tailored to deal with observed behavior that involves object manipulation, while Konidairs's method is more tailored to deal with observed behavior that involves navigation in an environment.

Given single observed behavior on stacking two blocks, we demonstrate that our method can successfully reproduce the stacking behavior on a physical robot. But the robot doesn't fully understand the general "stacking" concept from a single observed behavior, it only learns to stack two blocks with the relative pose between those two blocks being approximately the same with the one that was observed.

For the robot to take the learning target for imitation as learning to stack two blocks in a more general sense instead of learning to stack two blocks with a specific relative pose between those blocks, one of the suggested ways is to provide multiple demonstrations on stacking two blocks with varying relative poses. As discussed in \cite{abdo2013learning}, when multiple demonstrations are provided, the robot will be able to analyze what are the desired end-effects, i.e., the learning target of imitation, and conclude that the relative pose between those blocks do not need to be exact same as the one observed, but the blue block should be sitting on the green block.

\section{Conclusion}

We presented a general framework to learn policies for object manipulation by imitating an observed behavior. We show that based on our framework, given an observed behavior, we can segment the behavior into multiple skills if needed, and select the appropriate state abstraction such that 1) the relevant objects are identified, and 2) correct object-centred reference frame is selected. And our physical robot can learn policies that reproduce the observed behavior given RL problems formulated based on the segmented behavior and selected state abstractions.

Furthermore, we show that when private information is not available in the observed behavior, the robot is able to decide whether the private information is important, and reformulate the RL problems if needed. Specifically, in our experiment, our physical robot Baxter was able to recognize that the gripper action is important for carrying the block around and stacking it onto another block, and then it learns to 1) grasp the block with a proper holding force after its end-effector is at a grasp pose w.r.t the block, and 2) release the block by opening its gripper when the stacking is complete so that it can retrieve its end-effector.

\bibliographystyle{plain}
\bibliography{mybib}

\clearpage
\begin{appendices}
\section{Visual Processing} \label{app:vision}
It is assumed that the human demonstration and robot imitations involve manipulations on a stable tabletop. And only the objects that are on the table will be taken into account. Thus we just need to identify the tabletop, the objects on the table, and the hand at the starting frame, then track the objects and the hand. All the vision processing is built on PCL \cite{Rusu_ICRA2011_PCL} and OpenCV library \cite{opencv_library}.

\begin{figure*}[h!]
\centering
\subfigure[]{
\includegraphics[height=0.33\textwidth, width=0.48\textwidth]{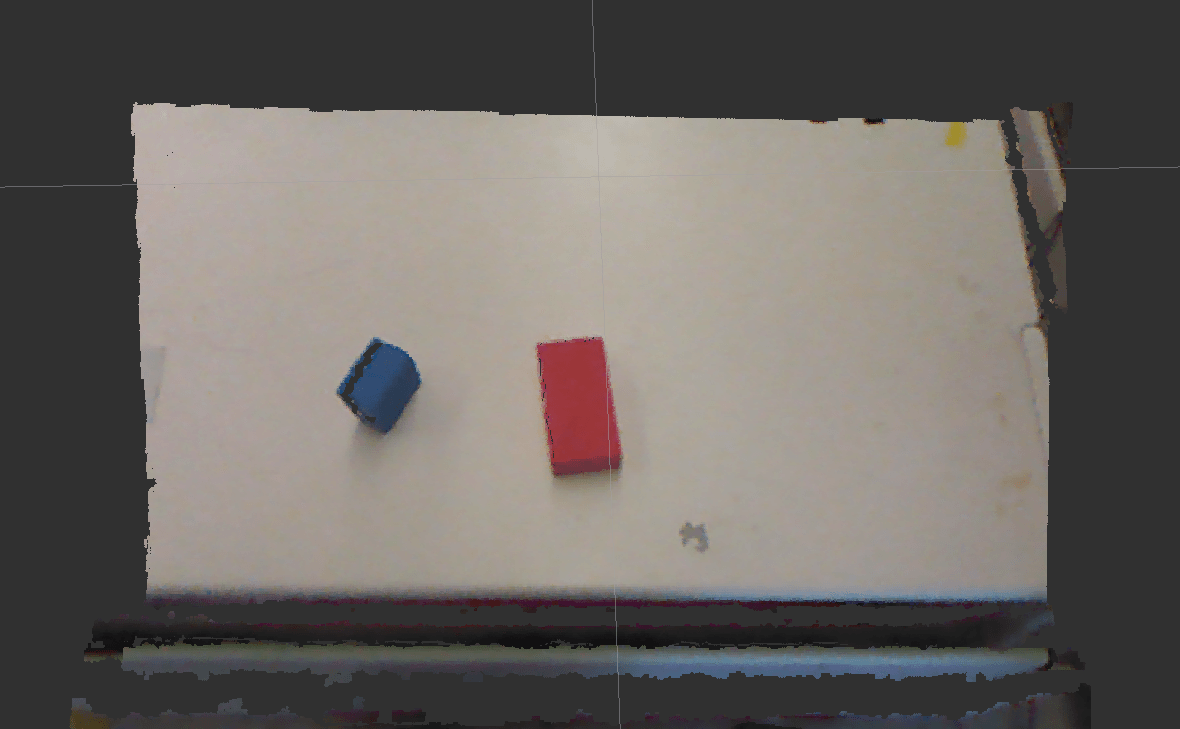} 
}
\subfigure[]{
\includegraphics[height=0.33\textwidth, width=0.48\textwidth]{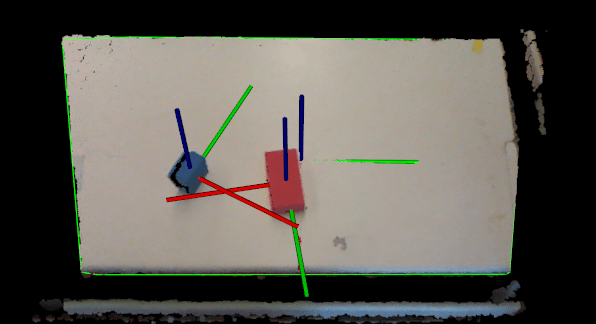} 
}
\caption{(a)Point cloud input from Kinect. (b)localized tabletop and blocks.\label{fig:segmentation}}
\end{figure*}

Given the starting frame of the RGB-D video, the tabletop is first identified.  A 3D plane is fitted to the fetched 3D point cloud data, and with the assumption that the dominant plane correspond to the tabletop, the tabeltop plane is easily detected. After clustering all the points that belong to the plane, the largest cluster is identified as the tabletop. The tabletop reference frame is localized as shown in Figure \ref{fig:segmentation}.

After identifying the tabletop, the 3D points above the tabletop are clustered into groups based on euclidean distance. Each cluster correspond to one object. To get the length and the width of a block, we first project the 3D points of the block cluster onto the localized tabletop plane, and then we find a rotated rectangle of the minimum area enclosing the projected 2D points, the length and the width of the rectangle is directly the length and the width of the block. The block height is calculated as the average of the heights of the three highest  3D points belong to the block cluster. The block pose is defined by placing the origin at the geometric center of the block model, and orienting the {\em x-axis} along the width of the block, orienting the {\em y-axis} along the length of the block, and orienting the {\em z-axis} along the height of the block, as shown in Figure \ref{fig:segmentation}. 

Note that the human hand is not easy to localize and track since it is not rigid and it is hard to obtain a good model, thus we focus our attention on marked cylinders (Figure \ref{fig:demo}) to simplify the problem. The left and right fingers represent the hand. In the experiments, the fingers are identified and tracked by color thresholding. The left and right finger 3D locations are the 3D locations of the centroids of the localized left and right finger cylinders. The human hand pose is defined by orienting the {\em z-axis} along the average of the axes of the cylinders, and orienting the {\em y-axis} along the vector pointing from the left finger centroid to the right finger centroid, and {\em x-axis} can be calculated from the cross product of the other two axes. The origin of the human hand pose is placed at a 3D point resulting from shifting the average 3D centroid of the two fingers along the {\em z-axis} away from the tabletop by around 10cm, as shown in Figure \ref{fig:demo}.

For robot imitations, the gripper pose in the robot body frame is directly available through the robot state publisher, and the robot knows (1)the transformation between the robot body frame and the camera coordinate frame(because of calibration), (2)the transformation between the camera coordinate frame and the tabletop coordinate frame, the robot can calculate the gripper pose in the tabletop coordinate frame.

The block is tracked using particle filters based on the geometry of the block. For more details, please refer to the technical report \cite{zhenz_tracker_math_2015}.

\section{Original DMP Formulation V.S. Biologically-inspired DMP Formulation} \label{app:dmp_vs}
Before getting to know the difference between the original DMP formulation \cite{schaal2006dynamic} and the biologically-inspired DMP formulation \cite{hoffmann2009biologically}, it is helpful to first go through a thorough introduction on DMP, available at \cite{zhenz_dmpmath_2015}.

\begin{figure}[h]
\centering
\centerline{\includegraphics[width=1.3\textwidth]{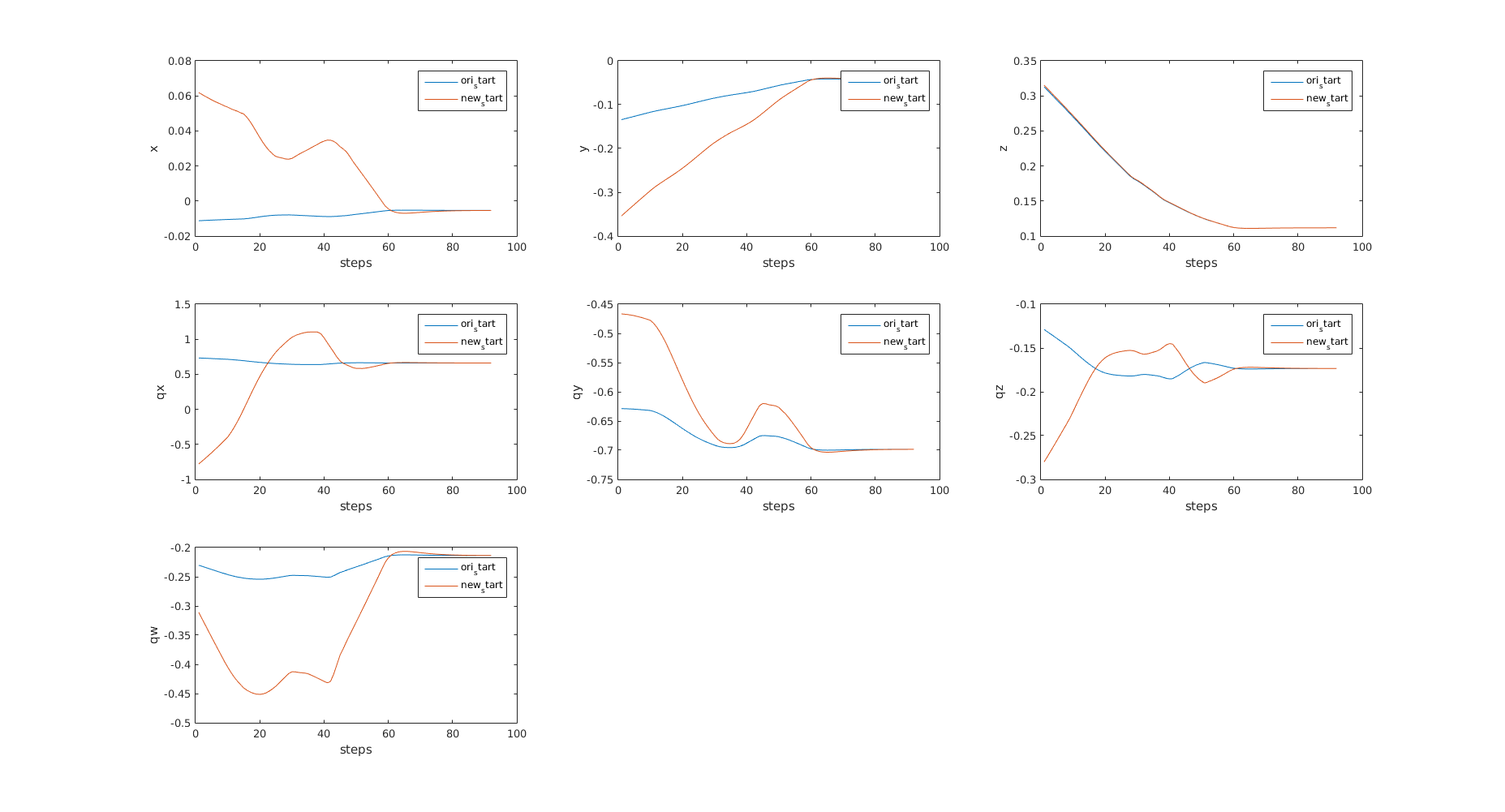}}
\caption{Trajectory of each dimension in gripper pose based on original DMP formulation \cite{schaal2006dynamic}. From left to right, top down are trajectory of: $x,y,z,qx,qy,qz,qw$ which represent robot's end-effector position and orientation respectively. \textit{Blue}: Trajectory generated with the observed start pose; \textit{Red}: Trajectory generated with a new start pose.}
\label{fig:dcp}
\end{figure}

\begin{figure}[h!]
\centering
\centerline{\includegraphics[width=1.3\textwidth]{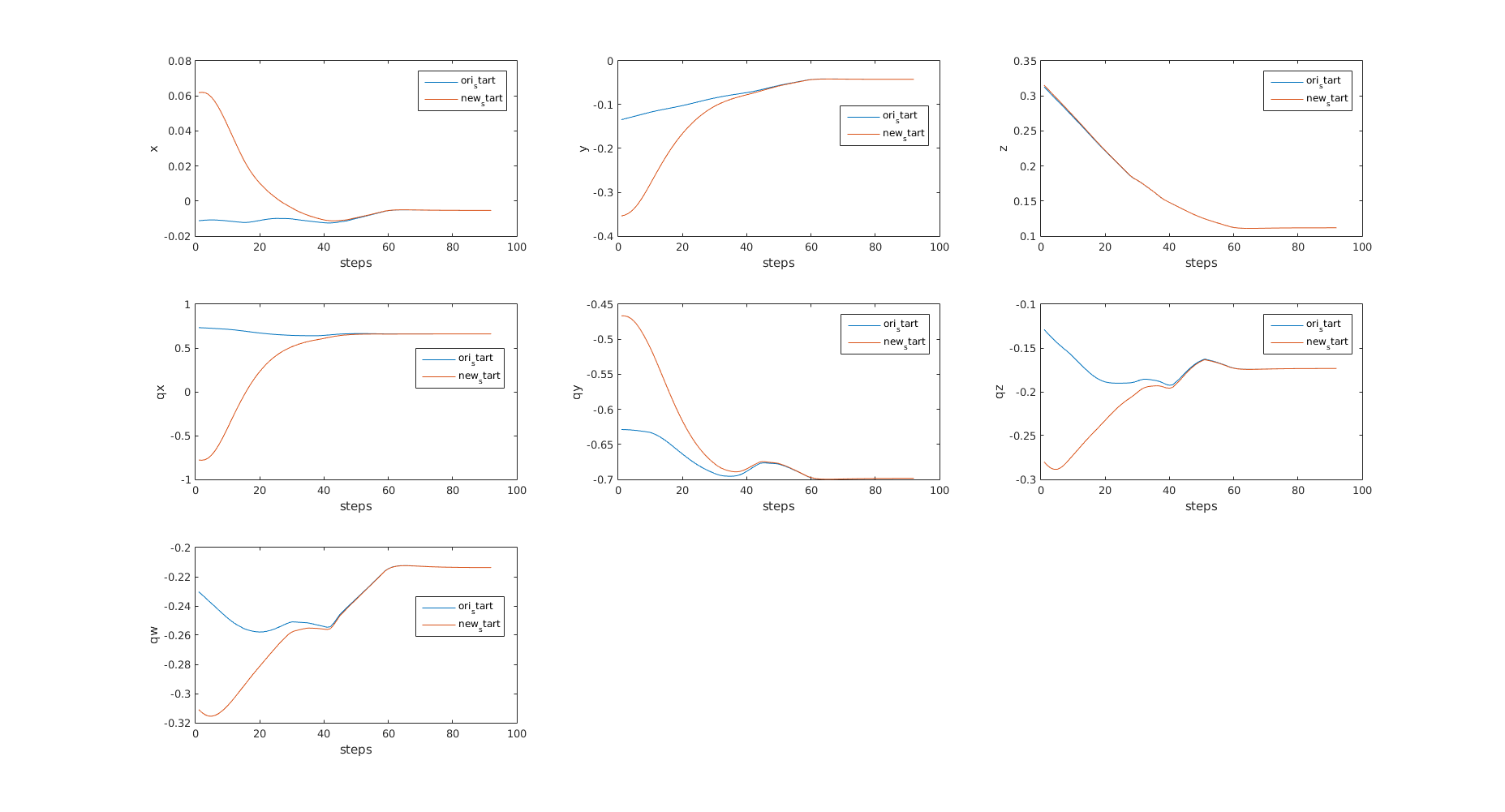}}
\caption{Trajectory of each dimension in gripper pose based on biologically-inspired DMP formulation \cite{hoffmann2009biologically}. From left to right, top down are trajectory of: $x,y,z,qx,qy,qz,qw$ which represent robot's end-effector position and orientation respectively. \textit{Blue}: Trajectory generated with the observed start pose; \textit{Red}: Trajectory generated with a new start pose.}
\label{fig:dcp_bio}
\end{figure}

As pointed out by Hoffmann et al. \cite{hoffmann2009biologically}, the original DMP formulation has some problematic shortcomings:
\begin{itemize}
\item If the goal $g$ is close to the start point $y_0$, a small change in $g$ may lead to huge accelerations that break the limits of the robot. As illustrated in the middle plot of trajectory of quaternion $y$ value in Figure \ref{fig:dcp}. 
\item If changing $g$ across the zero point, the whole movement inverts.  As illustrated in the middle left and right plot of trajectory of quaternion $x,z$ value in Figure \ref{fig:dcp}. 
\end{itemize}
The trajectory generated based on the modified DMP formulation is as shown in Figure \ref{fig:dcp_bio}. As we can see, both shortcomings as mentioned above are overcame.

We expect the biologically-inspired DMP formulation to suit our needs better, i.e., it adapts to new starting gripper pose better than the original DMP formulation, and this is confirmed in the experiments on both physical and simulated robots. In both experiments, the robot is first given a sample trajectory for reaching to grasp a block at a specific position, and then the block is placed at several new locations(and orientations) for the robot to reach, and we'll evaluate the performance of the DMP formulation based on how many new block poses can it adapt to such that the robot can successfully reach to the block with a grasp pose.

Our physical experiment is carried on our Baxter robot, the block is placed at 5 different locations(and orientation) for Baxter to reach. The experiment is as recorded in online videos \url{https://youtu.be/YpIpBgUuVqQ} and \url{https://youtu.be/P0XvoZtDZeU}. As we can see, based on the original DMP formulation, Baxter was only able to adapt its movements to 2 new starting poses; while based on the biologically-inspired DMP formulation, Baxter was able to adapt its movements to 3 new starting poses.

\begin{figure*}[h!]
\centering
\subfigure[]{
\includegraphics[height=0.40\textwidth, width=0.48\textwidth]{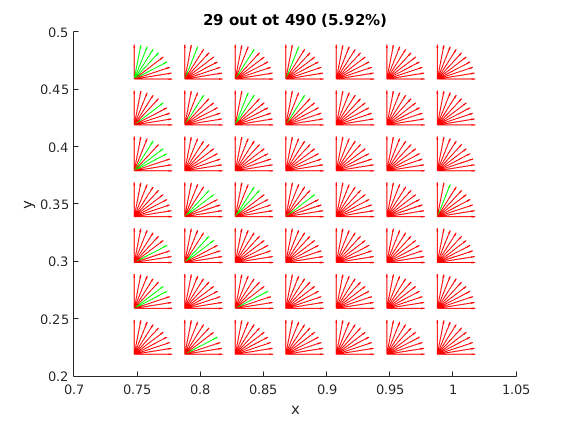} 
}
\subfigure[]{
\includegraphics[height=0.40\textwidth, width=0.48\textwidth]{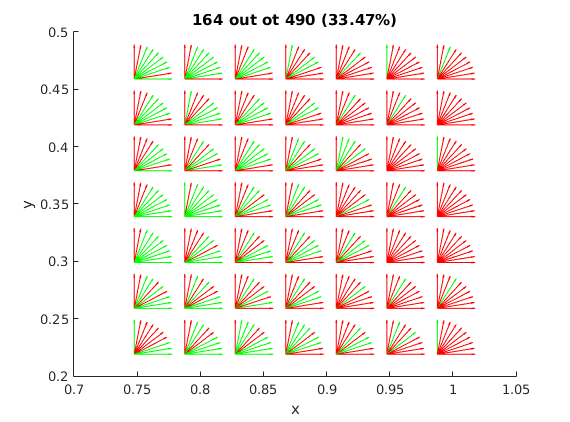} 
}
\caption{(a)Performance based on original DMP formulation. (b)Performance based on biologically-inspired DMP formulation. The {\em x-axis} and {\em y-axis} refers to the position of the block center w.r.t Baxter body frame. The block is placed on a table in front of Baxter, thus the $z$ locations of different test block poses are the same, so we are not showing it in the plots. The arrows indicates how the dominant axis of the block is oriented in Baxter body frame. \textit{Red arrow}: fail to reach to block at a grasp pose; \textit{Green arrow}: succeed to reach to block at a grasp pose.  \label{fig:simulation_vs}}
\end{figure*}

To conduct the experiment in a more systematic and efficient way, we test the performance of these two DMP formulations in simulation(Gazebo), and sample a set of block poses by drawing grids on location and orientations as the test set, and the test sample size is 490, far more than 5 in the physical experiment. The performance is as shown in Figure \ref{fig:simulation_vs}. Based on the original DMP formulation, Baxter only successfully adapt to 29 new block poses out of 490 test poses, achieving performance rate at 5.92\%. On the other hand, based on the biologically-inspired DMP formulation, Baxter can successfully adapt to 164 new block poses out of 490 test poses, achieving a much higher performance rate at 33.47\%, which is a reasonable performance given that only one sample trajectory is provided for a specific block pose.

To learn to adapt end-effector trajectory for more generalized target poses from a machine learning perspective, please refer to \cite{ude2010task}. In our work, we are relying on reinforcement learning to automatically adjust the end-effector trajectory for new block poses that are difficult to adapt to purely based on DMP formulation. 

\section{PoWER V.S. PI$^2$}\label{app:power_vs_pi2}
We are going compare two promising RL algorithms(policy search methods): Policy Learning by Weighting Exploration with the Returns(PoWER)\cite{kober2009policy}\cite{kober2009learning}, and Policy Improvement with Path Integrals(PI$^2$)\cite{theodorou2010reinforcement}\cite{theodorou2010generalized}
. Both algorithms can improve policy parametrized as a Dynamic Movement Primitive(DMP)\cite{ijspeert2013dynamical}.

\begin{figure}[h]
\centering
\includegraphics[width=1.0\textwidth]{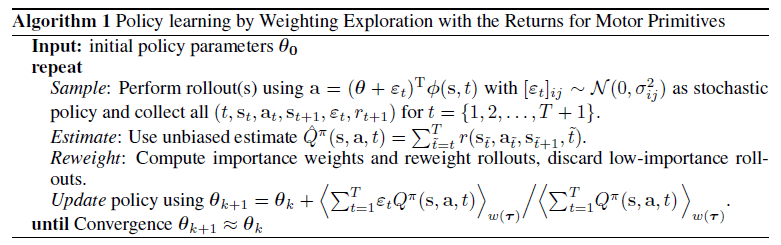}
\caption{PoWER algorithm\cite{kober2009policy}.}
\label{fig:power_alg}
\end{figure}

PoWER is an EM-inspired probabilistic policy improvement method. The PoWER algorithm is as shown in Figure \ref{fig:power_alg}. The policy parameters $\boldsymbol{\theta}$ is equivalently the weight vector $\mathbf{w}$ in DMP formulation as mentioned in sec \ref{sec:policy_learning}, and action $a$ corresponds to $f$ in DMP formulation in sec \ref{sec:policy_learning}. For more details about the algorithm, please refer to \cite{kober2009policy}\cite{kober2009learning}. PI$^2$ transforms policy improvements into an approximation problem of a path integral. The pseudocode of the algorithm is as shown in Figure \ref{fig:pi2_alg}. For more details about the algorithm, please refer to \cite{theodorou2010reinforcement}\cite{theodorou2010generalized}.

\subsection{Experiment Setup}
To compare the performance of these two algorithms, we focus on a reaching task in 1D. The reaching trajectory is generated by a DMP, and we prefer a trajectory with small accelerations(since we would like for Baxter to move gracefully), thus we will apply both policy search methods and choose the one that converges faster to policy parameters that generates small accelerations.

Specifically, the starting point of the trajectory is $y_0=0$, and the goal is $g=1$. The duration of the DMP is 1 second. 10 basis functions $psi_i(x)$ (Gaussian kernels) are used, with centers $c_i$ equally spaced in time(which corresponds to an exponential spacing in $x$), and variance $\sigma_i=\frac{1}{2}(c_i - c_{i-1})$. And other constants are $\alpha_z=25, \beta_z=\alpha_z/4, \alpha_x=\alpha_z/3$.

With the settings introduced above, the trajectory generated by the damped spring model without the forcing term $f$ is as shown in Figure \ref{fig:trajectory} in red.

\subsection{Rules of Reward(or Cost) Function Design}
As pointed out in \cite{theodorou2010generalized}, the immediate rewards in PoWER need to behave like an improper probability, that is, be strictly positive
and integrate to a constant number—this property can make the design of suitable cost functions
more complicated. Usually exponential function is used.

For PI$^2$, in contrast, in the immediate cost function as shown in Figure \ref{fig:pi2_alg}, $q_t=q(\mathbf{x_t})$ can be arbitrary state-dependent cost function, and $\mathbf{R}$ needs to be positive semi-definite. Thus it is generally easier to design cost function for PI$^2$ compared to PoWER.

With the rules above in mind, and to learn a policy that generates small acceleration, we design the immediate reward(or cost) function to be
\begin{align}
PoWER: r_t &=\frac{1}{n}e^{-0.01|\ddot{y_t}|}
\label{eq:reward}
\\
PI^2: r_t &=q_t = \frac{1}{n}(1-e^{-0.01|\ddot{y_t}|}) \ \ \  (\mathbf{R}=0)
\label{eq:cost}
\end{align}\label{eq:rewardcost}
where $\ddot{y}$ is the acceleration, and $n$ is the length of the trajectory(we say a trajectory $(y_0, y_1, \cdots, y_{n-1})$ has length $n$).

Note that for PoWER, $r_t$ is immediate reward that we would like to maximize, on the other hand, for PI$^2$, $r_t$ is immediate cost that we would like to minimize, and that explains why there is an additional "1-" in the $r_t$ for PI$^2$.

\subsection{Results}

\begin{figure*}
\centering
\subfigure[]{\label{fig:trajectory}
\includegraphics[height=0.40\textwidth, width=0.48\textwidth]{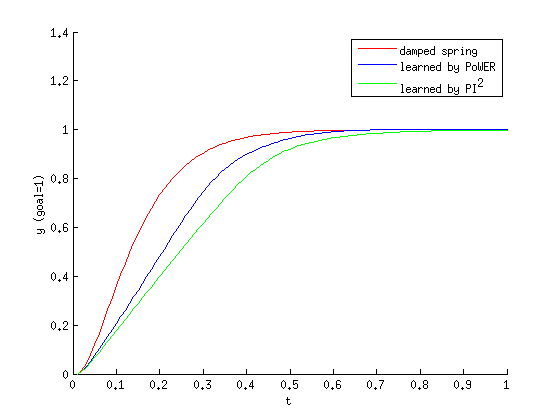} 
}
\subfigure[]{\label{fig:score}
\includegraphics[height=0.40\textwidth, width=0.48\textwidth]{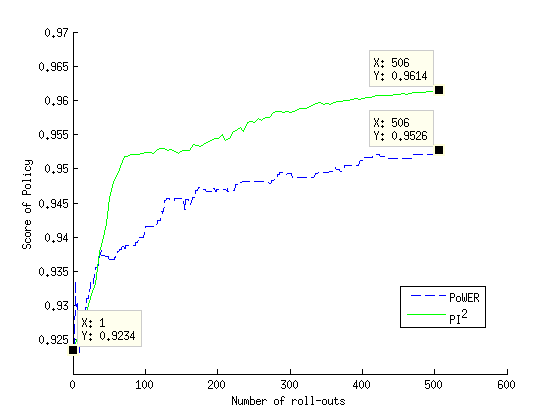} 
}
\caption{(a)\textit{Red}: reaching trajectory (goal $g$=1) generated by damped spring model without forcing term $f$; \textit{Blue}: reaching trajectory generated with policy learned by PoWER after 505 roll-outs;\textit{Green}: reaching trajectory generated with policy learned by PI$^2$ after 505 roll-outs. (b)\textit{Blue}: score of policy updated by PoWER;\textit{Green}: score of policy updated by PI$^2$.}
\end{figure*}

We apply PoWER and PI$^2$ with the immediate reward(or cost) function \ref{eq:reward}, \ref{eq:cost} designated for each method. In both cases, the initial policy parameters $\mathbf{w}=0$, which corresponds to pure damped spring model.

For each algorithm, we ran 505 roll-outs, and we evaluate the policy every time the policy parameters $\mathbf{w}$ are updated. The score of a given policy is evaluated as 
\begin{equation}
S(\mathbf{w}) =\sum_{t=0}^{n-1}e^{-0.01|\ddot{y_t}|}
\end{equation}
where $(\ddot{y}_0, \ddot{y}_1, \cdots, \ddot{y}_{n-1})$ is the acceleration profile generated by policy arametrized by $\mathbf{w}$ at the evaluation time. 

The score of the policy improves as the policy gets updated through roll-outs, as shown in Figure \ref{fig:score}, both algorithms start with the same initial policy evaluated as 0.9234, and after 505 roll-outs, PoWER ended up with a final policy evaluated as 0.9526, while \textbf{PI$^2$} ended up with a final policy evaluated \textbf{higher} at 0.9614.

As shown in Figure \ref{fig:trajectory}, the trajectory generated by the final policy learned by PoWER generate is steeper than the one generated by the final policy learned by PI$^2$, which also indicates that PI$^2$ ended up with a better policy that generates smaller acceleration.

\subsection{Decision on Using PI$^2$}
In the experiment of reaching task, PI$^2$ performs better than PoWER in that it learns a final policy that generates smaller acceleration.

PI$^2$ also offers the option to include the goal pose parameters into the policy paramerization, so the robot can learn not only the shape of the end-effector trajectory for manipulation tasks, but also the goal pose for manipulation tasks to deal with object location uncertainty due to perception noise, as shown shown in our Baxter experiment discussed in online presentation \cite{zhenz_pi2_2015}.

Since most of the end-effector trajectories of Baxter that we are going to deal with are similar to reaching, our experiment shows that PI$^2$ performs better than PoWER in learning a DMP for reaching, and it is generally easier to design cost function for PI$^2$ compared to PoWER, plus that PI$^2$ has the ability to incorporate learning of goal pose parameters, thus we conclude that PI$^2$ suits us better than PoWER.

\end{appendices}

\end{document}